\documentclass[final,5p,twocolumn,times]{elsarticle}
\usepackage{graphicx} 
\usepackage{graphics} 
\usepackage{times} 
\usepackage[T1]{fontenc} 
\usepackage{textcomp}   

\usepackage{enumitem}
\usepackage{amsthm}
\newtheorem*{remark}{Remark}
\newtheorem{thm}{Theorem}
\newtheorem{lem}{Lemma}
\newtheorem{assumption}{Assumption}

\usepackage{amssymb}  
\usepackage[linesnumbered,ruled,vlined]{algorithm2e}
\SetKwInput{KwInput}{Input}
\SetKwInput{KwOutput}{Output}

\newcommand{\bmu}{{\boldsymbol{\mu}}}

\newcommand{\cN}{\mathcal{N}}

\newcommand{\bK}{\mathbf{K}}

\newcommand{\bx}{\mathbf{x}}

\newcommand{\mE}{\mathbb{E}}
\newcommand{\mP}{{\mathbb P}}
\newcommand{\mQ}{{\mathbb Q}}

\usepackage{hyperref}
\hypersetup{
    colorlinks=true,
    linkcolor=blue,
    filecolor=magenta,      
    urlcolor=cyan,
    pdftitle={Overleaf Example},
    pdfpagemode=FullScreen,
}
    
\usepackage{subcaption} 
\usepackage{color}

\usepackage{mathtools}

\begin{document}

\begin{frontmatter}


\title{Efficient Iterative Proximal Variational Inference Motion Planning
}

\fntext[equal]{These authors contributed equally to this work.}

\author[ece]{Zinuo~Chang\fnref{equal}}
\author[ae]{Hongzhe~Yu\corref{cor1}\fnref{equal}}
\ead{hyu419@gatech.edu}
\cortext[cor1]{Corresponding author.}

\author[ece]{Patricio~Vela}
\author[ae]{Yongxin~Chen}

\address[ece]{School of Electrical and Computer Engineering, Georgia Institute of Technology, Atlanta, GA}
\address[ae]{School of Aerospace Engineering, Georgia Institute of Technology, Atlanta, GA}

\begin{abstract}
We cast motion planning under uncertainty as a stochastic optimal control problem, where the optimal posterior distribution has an explicit form. To approximate this posterior, this work frames an optimization problem in the space of Gaussian distributions by solving a Variational Inference (VI) in the path distribution space. For linear-Gaussian stochastic dynamics, a proximal algorithm is proposed to solve for an optimal Gaussian proposal iteratively. The computational bottleneck is evaluating the gradients with respect to the proposal over a dense trajectory. To tackle this issue, the sparse planning factor graph and Gaussian Belief Propagation (GBP) are exploited, allowing for parallel computation of these gradients on Graphics Processing Units (GPUs). We term the novel paradigm the \textit{Parallel Gaussian Variational Inference Motion Planning (P-GVIMP)}.
Building on the efficient algorithm for linear Gaussian systems, we then propose an iterative paradigm based on Statistical Linear Regression (SLR) techniques to solve planning problems for nonlinear stochastic systems, where the P-GVIMP serves as a sub-routine for the linearized time-varying system at each iteration. The proposed framework is validated on various robotic systems, demonstrating significant speed acceleration achieved by leveraging parallel computation and successful planning solutions for nonlinear systems under uncertainty. An open-sourced implementation is presented at \href{https://github.com/hzyu17/VIMP}{https://github.com/hzyu17/VIMP}.
\end{abstract}

\end{frontmatter}

\section{Introduction}
\label{sec:intro}

Motion Planning is one core decision-making component in autonomous robotic systems \cite{lavalle2006planning, gonzalez2015review}. Given an environment, a start configuration and a goal configuration, a motion planner computes a trajectory connecting the two configurations. The trajectory optimization paradigm \cite{ratliff2009chomp, schulman2014motion} formulates the motion planning problem as an optimization over all admissible trajectories with dynamics and environment constraints. An `optimal' plan is obtained by solving an optimization program that minimizes specific optimality criteria, such as time consumption and control energy.

Uncertainties such as sensing \cite{chen2022should}, actuation noises, and external disturbances \cite{chen2000nonlinear} arise and affect the quality of the motion plans in the execution phase. To reduce their impacts, motion planning under uncertainties takes into account the uncertainties in the planning phase in their formulations \cite{kalakrishnan2011stomp}. Stochastic optimal control and probabilistic robotics \cite{aastrom2012introduction, thrun2002probabilistic, hoshino2025path} provide a principled framework to address this problem, where noise is explicitly modeled in robot dynamics, and a statistical optimality index is minimized over the trajectory distribution space. 

Gaussian Process Motion Planning (GPMP) paradigm \cite{mukadam2016gaussian, mukadam2018continuous} used the linear SDE dynamics model to formulate motion planning problems as probability inference, where the optimal solution is formulated as a posterior probability. The objective in GPMP is to solve a trajectory that maximizes this posterior. 

The solution obtained from GPMP is still deterministic. Variational Inference (VI) \cite{hoffman2013stochastic, blei2017variational} formulates the inference problem into an optimization by minimizing the Kullback–Leibler (KL) divergence between a proposal distribution and the target posterior. VI has been applied to solve for a trajectory distribution in the planning-as-inference formulation \cite{yu2023gaussian, yu2023stochastic, cosier2024unifying, power2024constrained}. Gaussian Variational Inference Motion Planning (GVIMP) \cite{yu2023gaussian, yu2023stochastic} showed satisfying performance in challenging tasks such as planning through narrow gaps \cite{hsu1998finding}. The GVIMP formulation is related to entropy-maximized motion planning and reinforcement learning \cite{chen2016relation, chen2016optimal, haarnoja2018soft, zhao2019maximum}. 

Introducing distributional variables increases the dimension of the problem. Fortunately, in motion planning problems, the underlying probabilistic graph enjoys a sparsity pattern \cite{barfoot2014batch, mukadam2018continuous} which is leveraged in algorithm designs to factorize the target posterior and reduce the time complexity \cite{yu2023stochastic, barfoot2020exactly}. In this paper, we propose the proximal point GVI algorithm \cite{khan2015kullback, khan2015faster} over the motion-planning factor graph. The iterative update rule is presented in an explicit form, and its computations are naturally distributed across the factor levels of the graph.

Similar to deterministic trajectory optimization paradigms \cite{schulman2014motion, mukadam2018continuous}, the collision-checking step of the algorithm on a dense trajectory is computationally demanding in GVIMP and represents the computational bottleneck \cite{yu2023stochastic}. The sparse factor graph is leveraged in this paper to distribute collision-checking computations in parallel on the GPU. We also deploy the Gaussian Belief Propagation (GBP) algorithm \cite{shental2008gaussian} for efficient marginal covariance computation.

Most existing planning-as-inference paradigms \cite{mukadam2018continuous, yu2023gaussian, osa2020multimodal, carvalho2023motion} assume linear stochastic differential equations (SDEs) as the underlying dynamics. Unlike these works, this study addresses \textit{nonlinear} SDEs \cite{sarkka2019applied}, under the standard Gaussian transition assumption widely adopted in robotics applications \cite{dellaert2021factor, barfoot2024state}. We iteratively apply Statistical Linear Regression (SLR) to obtain a linear time-varying (LTV) system at each time, and perform P-GVIMP on the resulting LTV system. The proposed algorithm integrates the sequential Bayesian principles \cite{seo2024sequential} into the GVIMP framework for motion planning tasks.

\textit{\textbf{Related Works.}}

{\em (a) Proximal Variational Inference.}
The proximal point algorithm was first proposed in \cite{khan2015kullback}, where the KL-divergence was used as the proximal operator for distributions. The connection between the proximal-point and natural gradient descent algorithms was pointed out in \cite{blei2017variational}. \cite{khan2015faster} generalized the proximal point variational inference by replacing the proximal term with general divergence functions. Recently, \cite{diao2023forward} proposed a proximal gradient algorithm by decomposing the KL-divergence objective into a potential and an entropy term. The above algorithms were applied directly to the posteriors without considering their factor graph structure.

{\em (b) Approximations for nonlinear stochastic processes.}
The iterative linearization techniques for nonlinear systems approximate a nonlinear model with a linear one at each time step. The Extended Kalman Filter (EKF)~\cite{gelb1974applied} uses a single Taylor series expansion. Statistical linearization techniques for nonlinear systems in filtering, such as the Unscented Kalman Filter (UKF)~\cite{julier2004unscented}, offer a powerful alternative to analytical methods like the EKF. These methods, often referred to as sigma-point filters, deterministically sample points from the state's probability distribution and then propagate these points through the nonlinear function. The mean and covariance of the transformed points are then used to create a linearized model. This approach doesn't require computing derivatives and can more accurately capture the statistics of the transformed distribution, especially for highly nonlinear systems. To the best of the author's knowledge, our work is the first to use the sigma-point paradigm to solve nonlinear \textbf{planning} problems.

{\em (c) Parallel Motion Planning.}
Parallel motion planning has seen concrete GPU and multi-core implementations that significantly accelerate planning. \cite{pan2012collision} and \cite{pan2012gpu} demonstrated GPU-accelerated collision checking and PRM construction, achieving orders of magnitude speedups. \cite{janson2015fast} and \cite{gammell2015batch} exploited batch operations to parallelize sampling-based planning. In trajectory optimization, STOMP~\cite{kalakrishnan2011stomp} explicitly evaluated multiple rollouts in parallel. For multi-agent systems, ORCA~\cite{van2011reciprocal} and parallelized CBS variants~\cite{sharon2015conflict} enabled scalable multi-robot coordination. Finally, GPU-RRT-Connect~\cite{huang2025prrtc} extended parallelism to the full RRT pipeline, showing consistent speedups over CPU baselines. All previous studies addressed parallel collision checking for deterministic motion planning, whereas this paper considers collision checking in a probabilistic setting by computing the expected collision cost with respect to a proposal distribution over a dense trajectory.

{\textbf{Key Contributions.}} The contributions of this research are: (1) We propose a KL-proximal variational inference algorithm for robot motion planning and show that its proximal gradient computations parallelize naturally by exploiting the splitting structure and the underlying sparse factor graph; (2) A parallel KL-proximal algorithms is proposed via computation implementation on GPUs, where parallel collision cost computation and Gaussian Belief Propagation are used to compute marginal factors efficiently; (3) We propose a sequential statistical linear regression scheme for motion planning of nonlinear dynamical systems, where the parallel KL-proximal algorithm is applied to the linearized systems as a sub-module at each outer iteration. 

Planning results are demonstrated on various robot models, and comparison studies have been conducted to demonstrate the efficiency improvement. This paper builds upon the authors' prior studies \cite{yu2023gaussian, yu2023stochastic} by addressing computational efficiency, exploring the KL-proximal GVI algorithm in the path-distribution space for motion planning, and demonstrating the applicability of the GVIMP approach to nonlinear dynamics.

This paper is organized as follows. Section \ref{sec:Preliminaries} introduces the background knowledge. Section \ref{sec:problem_formulation} introduces the KL-proximal GVI algorithm and its structure for planning problems. Section \ref{sec:method} introduces the sparse factor graph and marginal gradient updates for the KL-proximal algorithm. Section \ref{sec:p-gvimp} introduces the iterative P-GVIMP algorithm for nonlinear systems. The proposed framework is illustrated in Section \ref{sec:experiments} through numerical experiments, followed by a conclusion in Section \ref{sec:conclusion}.

\section{Variational Inference Motion Planning (VIMP)}
\label{sec:Preliminaries}
This section introduces some important preliminary knowledge of this paper, including motion planning under uncertainty and variational inference motion planning.
\subsection{Planning under Uncertainties as Stochastic Control}
\label{sec:planning_as_stochastic_control}
We address motion planning under uncertainty within a stochastic control framework. For the \textit{nonlinear} process
\begin{equation}\label{eq:nonlinear_dynamics}
    dX_t = f(X_t, u_t)\,dt + g_t(X_t)\,dW_t,
\end{equation}
The objective to be minimized is
\begin{align}
\label{eq:SCP_formulation}
    \!\!\!\! \min_{X(\cdot), u(\cdot)}& \mathcal{J} \triangleq \mE \left \{\int_{0}^{T} \frac{1}{2}\|u_t\|^2 + V(X_t) d t + \Psi_T(X_T) \right \}
\end{align}
where the running cost consists of a state-related cost $\int_{0}^{T} V(X_t) d t$, a control energy cost $\int_{0}^{T} \frac{1}{2}\|u_t\|^2 d t$, and $\Psi_T(x_T)$ is a state terminal cost. The state costs regulate the desired behaviors, such as collision avoidance. The \textit{prior} process associated with the process \eqref{eq:nonlinear_dynamics} is defined by letting $u_t \equiv 0$, leading to
\begin{equation}
    \label{eq:dynamics_nonlinear_uncontrolled}
    dX_t = f_0(X_t) dt + g_t(X_t) dW_t.
\end{equation}

{\em Gaussian assumptions.}
The state distribution of system \eqref{eq:nonlinear_dynamics} follows the Fokker-Planck (FPK) equations, leading to non-Gaussian distributions for the states under the nonlinearity \eqref{eq:nonlinear_dynamics}. It is hard to solve the FPK for general nonlinearities~\cite{sarkka2019applied}, and Gaussian assumptions are one popular approximation assumption in robotics applications~\cite{barfoot2024state, dellaert2021factor}.

\subsection{Gaussian VI Motion Planning (GVIMP)}
This section covers the problem \eqref{eq:SCP_formulation} as a path-distribution control problem, and proposes a Gaussian VI solution to it.

\paragraph{Control-inference duality}
Denote the measure induced by the controlled linear process \eqref{eq:dynamics_linear} as $\mQ$, and the measure induced by the prior process \eqref{eq:dynamics_nonlinear_uncontrolled} as $\mP$. By Girsanov's theorem, the following objective is equivalent to \eqref{eq:SCP_formulation} \cite{yu2023stochastic, yu2024optimal}
\label{eq:obj_KL}
\begin{align}
     \!\!\! \mathcal{J} &= \mE_{\mQ} [ \log\frac{d\mQ}{d\mP}  + \mathcal{V} ] \propto \mE_{\mQ} [ \log d\mQ - \log \frac{e^{-\mathcal{V}} d\mP}{\mE_{\mP} \left[ e^{-\mathcal{V}} \right] } ]  \nonumber
     \\
     &= {\rm KL} \left( \mQ \parallel \mQ^\star \right),
    \label{eq:obj_measure_2}
\end{align}
where the cost-related functional $\mathcal{V} \triangleq \int_{0}^{T} V(X_t) dt  + \Psi_T(X_T)$, and the measure $\mQ^\star$ is defined as 
\begin{equation}
\label{eq:defn_dmP_star}
    d\mQ^\star \triangleq \frac{e^{-\mathcal{V}}}{\mE_{\mP} \left[ e^{-\mathcal{V}} \right]} d\mP \propto e^{-\mathcal{V}} d\mP.
\end{equation}
Path space Variational Inference is formulated as \cite{mukadam2016gaussian, yu2023stochastic}
\begin{equation}
    q^{\star} = \underset{q_\theta\in \mathcal{Q}}{\arg\min}\; {\rm KL} \left ( q_\theta \parallel \mQ^\star \right )
\label{eq:GVI_formulation}
\end{equation}
where
$\mathcal{Q} \triangleq \{ q_\theta: q_\theta \sim \mathcal{N}(\mu_{\theta}, \Sigma_{\theta}) \}$ consists of the parameterized proposal Gaussian distributions.

\paragraph{Linear-Gaussian System Case} When the stochastic process \eqref{eq:nonlinear_dynamics} reduces to linear time-varying stochastic system 
\begin{equation}\label{eq:dynamics_linear}
    dX_t = (A_t X_t + a_t + B_t u_t) dt + B_t dW_t,
\end{equation} 
the associated \textit{prior} process is
\begin{equation}
    \label{eq:dynamics_linear_uncontrolled}
    dX_t = (A_t X_t + a_t) dt + B_t dW_t.
\end{equation}
Consider the above KL-minimization problem \eqref{eq:obj_measure_2} with the time discretization $\mathbf{t}\triangleq \left[t_0, \dots, t_N\right],\; t_0 = 0, \; t_N=T$.
For linear Gaussian prior dynamics \eqref{eq:dynamics_linear_uncontrolled}, the discrete-time path distribution $d\mP$ over $\mathbf{t}$ is a Gaussian distribution $\bx \sim \mathcal{N}(\bmu|\mathbf{K})$. The discrete-time cost factor $e^{-\mathcal{V}}$ is defined as
\[
e^{-\mathcal{V}} = e^{-\int_0^T V(X_t) d t + \Psi_T} \approx e^{-\sum_{i=0}^{N-1} V(X_i)\times \Delta t + \Psi_T},
\]
then the un-normalized discrete-time optimal distribution to the problem \eqref{eq:defn_dmP_star} and is defined as 
\begin{equation*}
    \mQ^\star(\bx) \propto e^{-\frac{1}{2}\lVert \bx - \bmu \rVert_{\mathbf{K}^{-1}}^2 -\sum_{i=0}^N V(X_i)}\triangleq \Tilde{q}^\star(\bx).
\end{equation*}
Under this discretization, the inference problem \eqref{eq:GVI_formulation} becomes 
\begin{align}
\label{eq:GVI_formulation_linear}
\begin{split}
    \underset{\mu_{\theta}, \Sigma_{\theta}}{\min}\; \mE_{q_\theta \sim \mathcal{N}(\mu_\theta, \Sigma_\theta) }{\left[V(\bx)\right]} + {\rm KL} \left (q_\theta \parallel \mathcal{N}(\bmu, \bK)\right).
\end{split}
\end{align}

Define the negative log probability for the posterior as $\psi(\bx) \triangleq -\log \Tilde{q}^\star(\bx)$, and the VI objective is rewritten as
\begin{equation*}
    \mathcal{J}(q_\theta) = {\rm KL}\left( q_\theta(\bx) \parallel \Tilde{q}^\star(\bx) \right) = \mE[\psi(\bx)] - \mathcal{H}(q_\theta),
\end{equation*}
where $\mathcal{H}(q_\theta)$ denotes the entropy of the distribution $q_\theta$, which promotes the robustness of the motion plan solution. An additional temperature parameter, $\hat{T}$, is added to trade off this robustness, leading to the objective of temperature \cite{yu2023gaussian}
\begin{equation*}
    \mathcal{J}(q_\theta) = {\rm KL}\left( q_\theta(\bx) \parallel \Tilde{q}^\star(\bx) \right) = \mE[\psi(\bx)] - \hat{T} \mathcal{H}(q_\theta)
\end{equation*}

\section{KL-Proximal Algorithm For Linear-Gaussian Systems}
\label{sec:problem_formulation}
This section derives a proximal point algorithm for the GVIMP problem for linear systems~\eqref{eq:GVI_formulation_linear}. 
\subsection{Proximal Point Gaussian Variational Inference}
\label{sec:KL-proxmal_point}

{\em (a) KL-proximal point Variational Inference.} At each step, the proximal point, or proximal minimization algorithm in the Euclidean space, solves the following optimization problem  
\begin{equation}
    \label{eq:proximal_point_Euclidean}
    x_{k+1} = \arg\min_x \mathcal{J}(x) + \frac{1}{2\beta_k} \lVert x - x_k \rVert^2_2,
\end{equation}
where $\lVert x - x_k \rVert_2^2$ is a regularizer centered around the previous iterate $x_k$, and $\beta_k$ is the step size. The proximal point algorithm is more stable than vanilla gradient descent, since it corresponds to the backward Euler integration of the gradient flow \cite{parikh2014proximal}.

The KL divergence is a better metric candidate regularizer than the Euclidean $2$-norm for distributions. Replacing the $2$-norm regularizer in \eqref{eq:proximal_point_Euclidean} by the KL-divergence, the KL-proximal point algorithm \cite{chretien2000kullback} is obtained at each step 
\begin{equation}
\label{eq:KL_proximal}
\delta \theta_{k+1} = \arg\max_{\theta} \mathcal{L}\left( x, \theta \right) - \frac{1}{\beta_k}{\rm KL}\left( q(x\mid \theta) \parallel q(x\mid \theta_k) \right).
\end{equation}

{\em (b) Splitting in Variational Motion Planning.} Solving the proximal point iteration for general nonlinear Evidence Lower Bounds (ELBOs) is challenging \cite{khan2015faster}. A common situation in variational inference is that the ELBO splits into two parts, $- \mathcal{L} = \psi_e + \psi_d$, where the gradients of $\psi_e \triangleq \mE_{q_\theta}\left[ \log\left(f_e(x)\right) \right]$ are easy to compute, sometimes are directly known in closed form, e.g., for Gaussian priors and other conjugate models; and the computation of gradients of $\psi_d \triangleq \mE_{q_\theta}\left[ \log\left(f_d(x)\right) \right] $ is not direct. To obtain updates efficiently, the first-order approximations are performed on $\psi_d$. This decomposition is also known as `splitting'. 

Inspired by this, this paper derives the splitting structure for motion planning problems for the first time. We decompose the motion planning ELBO into two different parts 
\begin{align}
\begin{split}
\!\!\!\mathcal{L}(\bx, \theta) &= \mE_{q_\theta} \left[\log\Tilde{\mQ}^* \right] - \mE_{q_\theta} \left[ \log q_\theta \right]
\\
&= \!-\mE_{q_\theta} \left[ \log q_\theta + \frac{1}{2}\lVert \mathbf{x} - \bmu \rVert_{\mathbf{K}^{-1}}^2 \right] \!- \!\mE_{q_\theta} \left[ V(\bx) \right],
\end{split}
\end{align}
where we use the two functions $\psi_e, \psi_d$ to denote the two parts
\begin{subequations}
\label{eq:psi_easy_hard}
\begin{align}
    \psi_e(\bx,\theta) &\triangleq \mE_{q_\theta} \left[ \log q_\theta + \frac{1}{2}\lVert \mathbf{x} - \bmu \rVert_{\mathbf{K}^{-1}}^2 \right], \;
    \\
    \label{eq:psi_hard}
    \psi_d(\bx,\theta) &\triangleq \mE_{q_\theta} \left[ V(\bx) \right].
\end{align}
\end{subequations}
The function $\psi_e$ is the negative likelihood of the prior cost-related term and the entropy, which is a conjugate model. The term $\psi_d$ is the cost related to collision avoidance, which is non-conjugate. Let $q_e \sim \mathcal{N}(\bmu, \bK)$. Then $\psi_e$ equals the KL divergence between $q_\theta$ and $q_e$, up to a constant
\begin{align*}
    \psi_e(\bx, \theta)  \propto  \mE_{q_\theta} \left( \log q_\theta - \log q_e \right) = {\rm KL}\left( q_\theta \parallel q_e \right), 
\end{align*}
which indicates that the gradients of $\psi_e$ with respect to $(\mu_\theta, \Sigma_\theta^{-1})$ have closed-form. The non-conjugate model, $\psi_d$, is approximated to the first order. This leads to the following KL proximal gradient update at each iteration.

{\em (c) The KL-proximal gradients with splitting.}
After splitting, the update step of the KL-proximal algorithm is derived. By substituting the terms $\psi_e$ and $\psi_d$ from \eqref{eq:psi_easy_hard} into \eqref{eq:KL_proximal}, we obtain
\begin{align}
\label{eq:prox-KL-optimization}
&\theta_{k+1}=\underset{\theta}{\operatorname{argmin}} \; \mathcal{J}_{\rm ProxKL}
\\
&\triangleq \underset{\theta}{\operatorname{argmin}} \; \theta^{T} \left[\nabla \psi_d\left(\theta_{k}\right)\right]+\psi_e(\theta)+\frac{{\rm KL}\left(q_\theta(\mathbf{x}) \| q\left(\mathbf{x} \mid \theta_{k}\right)\right)}{\beta_{k}}. \nonumber
\end{align}

Taking the first-order gradients with respect to $(\mu_\theta, \Sigma_\theta^{-1})$, we obtain the KL-proximal updates for them. The update rule is summarized in Theorem \ref{thm:KLproximal_updates}. See \ref{sec:appendix_A} for the proof.
\begin{thm}
\label{thm:KLproximal_updates}
The KL-proximal updates for $(\mu_\theta, \Sigma_\theta^{-1})$ are 
\begin{subequations}
\label{eq:mu_Sigma_update}
\begin{align}
    &\!\!\! ( \bK^{-1} + \frac{\Sigma_{\theta_k}^{-1}}{\beta_k}  ) \mu_{\theta_{k+1}} = -\nabla_{\mu_\theta} \psi_d  + \bK^{-1}\bmu + \frac{\Sigma_{\theta_k}^{-1}}{\beta_k}\mu_{\theta_k},
    \\
    &\Sigma_{\theta_{k+1}}^{-1} = \frac{\beta_k}{\beta_k + 1} ( 2 \nabla_{\Sigma_{\theta}} \psi_d  + \bK^{-1} + \frac{\Sigma_{\theta_k}^{-1} }{\beta_k} ).
\end{align}
\end{subequations}
\end{thm}
\begin{remark}
This update rule is accurate in the sense that it directly approximates the expected term $\psi_d(\theta) = \mathbb{E}_{q(\mathbf{x} \mid \theta)}\!\left[\|\mathbf{h}(\mathbf{x})\|^{2}_{\Sigma_{\mathrm{obs}}}\right]$ to first order, rather than approximating the integrand inside the expectation. 
Under $L$-smooth assumption on $\psi_d$, the approximation error is of order $\mathcal{O}(\|\theta-\theta_k\|^2)$ and hence remains bounded. \ref{sec:pgvimp_analysis} presents the proof.
\end{remark}

{\em \textbf{Convergence of the algorithm.}} First assume that the ELBO $\mathcal{L}$ is continuous and admits a maximum. Based on \cite{khan2015faster}, the following assumptions are required: (i) $\psi_d$ is $L$-smooth;  (ii) $\psi_e$ is convex;  (iii) There exists $\alpha > 0$ satisfying the specific monotonicity condition. Under these assumptions, the proposed KL-Proximal update satisfies $\|\theta_{k+1} - \theta_k\| \to 0$ when $0 < \beta_k < \alpha/L$. \ref{sec:pgvimp_analysis} provides a detailed statement and proof.

{\em (d) Connection to Natural Gradient Descent (NGD).} The NGD updates can be integrated into proximal point algorithms, using the symmetric KL-divergence as the proximal regularizer \cite{blei2017variational}. Specifically, the NGD update step with unit step size is
\[
\delta \theta = G(\theta)^{-1} \nabla_\theta \mathcal{L}(x, \theta),
\]
where $G(\theta)\triangleq \mE_{q_\theta} \left[ \nabla_\theta \log q_\theta \left(\nabla_\theta\log q_\theta\right)^T \right]$ is the Fisher information matrix. It is the solution to the following proximal point algorithm with the first-order approximation of the ELBO 
\begin{align}
\label{eq:ngd_proximal}
\delta \theta &= \arg\min_{d\theta} \; d\theta^T \nabla_\theta \mathcal{L}(x,\theta) + d\theta^T G(\theta) d\theta
\\
&\!\!\!\!\!\!\! \approx \arg\min_{d\theta} d\theta^T \nabla_\theta \mathcal{L}(x,\theta) + {\rm KL}^{\rm sym}(q(x\mid \theta) \parallel q(x\mid \theta+d\theta)),\nonumber
\end{align}
where the symmetric KL divergence ${\rm KL}^{\rm sym}$ between $2$ distributions $q_1$ and $q_2$ is defined as
\[
{\rm KL}^{\rm sym}(q_1 \parallel q_2) \triangleq {\rm KL}(q_1 \parallel q_2) + {\rm KL}(q_2 \parallel q_1). 
\]
With step size \(\beta_k\), the NGD update replaces the KL proximal term in \eqref{eq:prox-KL-optimization} with its symmetrized counterpart.

\section{Sparse Factor Graph For Marginal Computation} 
\label{sec:method}
In the KL-proximal updates \eqref{eq:mu_Sigma_update}, computing the gradients $\nabla_{\mu_\theta}\psi_d$ and $\nabla_{\Sigma_\theta}\psi_d$ represents the computational bottleneck. This section introduces a paradigm that leverages the underlying sparse motion planning factor graph to distribute this computation onto the marginal levels.
\subsection{Sparse Motion Planning Factor Graph}
For clarity, denote $\psi(\bx) = V(\bx)$ in \eqref{eq:psi_hard}. With some algebraic calculation, the gradients of $\psi_d$ are given by
\begin{subequations}
\label{eq:gradients_psid}
    \begin{align}
    \label{eq:gradients_psid_mu}
    \nabla_{\mu_\theta} \psi_d (\theta) &= \Sigma_\theta^{-1}\mathbb{E}\left[\left(\mathbf{x}-\mu_{\theta}\right) \psi \right]
    \\
    \label{eq:gradients_psid_Sigma}
    \nabla_{\Sigma_\theta} \psi_d (\theta) &= -\frac{1}{2} \Sigma_\theta^{-1} \mathbb{E}[\psi]  + \frac{1}{2} \Sigma_\theta^{-1} \mathbb{E}\left[(\mathbf{x}-\mu_\theta)(\mathbf{x}-\mu_\theta)^T \psi \right] \Sigma_\theta^{-1},
\end{align}
\end{subequations}
which breaks down into calculating the expectations of $\psi$
\begin{align*}
    \mE\left[ \psi \right], \; \mE\left[ \left( \bx - \mu_\theta \right)\psi \right], \; \mE\left[ \left( \bx - \mu_\theta \right)\left( \bx - \mu_\theta \right)^T \psi \right].
\end{align*}

The motion planning factor graph is shown in the following Figure~\ref{fig:factor_graph}. It contains two types of factors: the $f_{i,i+1}(x_i, x_{i+1})$ represents the prior factor enforcing the kinodynamics constraints, and the $f_i(x_i)$ are the collision factors that encourage the trajectory to remain within obstacle-free regions.
\begin{figure}[th]
    \centering
    \includegraphics[width=\linewidth]{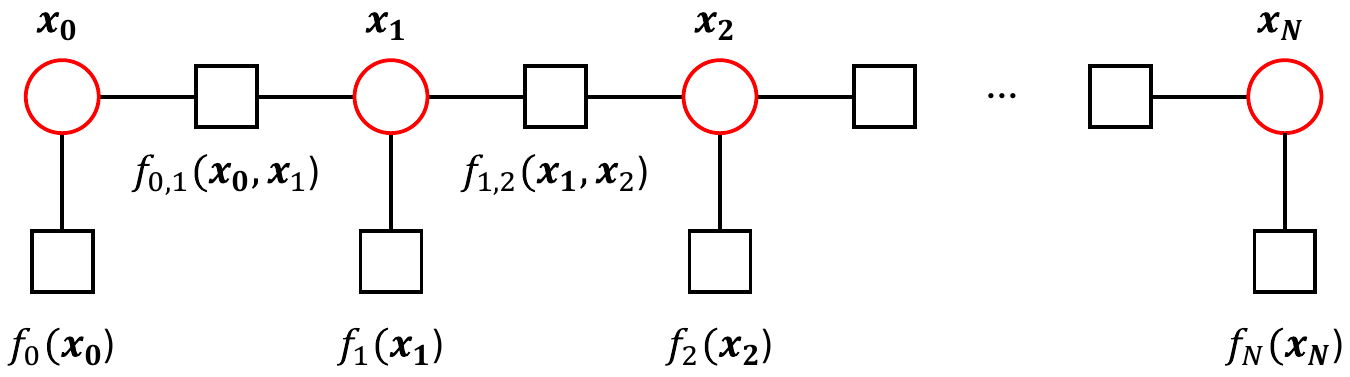}
    \caption{The GVIMP factor graph.}
    \label{fig:factor_graph}
\end{figure}

This factorized structure is leveraged to distribute the computation of $\psi_{d}$ on a GPU. 
Given the factor graph structure, we factorize the collision costs $\psi_d = \lVert h(x) \rVert^2_{\Sigma_{\rm obs}}$ into
\begin{subequations}
\begin{align*}
    \psi_d &= \psi^d_0(q_0), \dots, \psi^d_N(q_N),
\end{align*}
\end{subequations}
where the factorized potentials are defined as the collision cost on each state on the trajectory $\psi^{d}_i \triangleq \lVert h_i(q_i) \rVert_{\sigma_{\rm obs}}^{i}$.
The marginal Gaussian variables are linearly mapped from the joint Gaussian
\begin{equation}
\label{eq:mapping_from_joint_to_factor}
    q_{i} \sim \mathcal{N}\left(\mu_{\theta}^{i}, \Sigma_{\theta}^{i}\right), \quad \mu_{\theta}^{i}=M_{i} \mu_{\theta}, \quad \Sigma_{\theta}^{i}=M_{i} \Sigma_{\theta} M_{i}^{T}.
\end{equation}

\subsection{Marginal updates}
\label{sec:factorization}

By Stein's Lemma, \eqref{eq:gradients_psid_mu} and \eqref{eq:gradients_psid_Sigma} are written as
\begin{subequations}
    \begin{align}
    \label{eq:gradients_mu_stein}
    \nabla_{\mu_\theta} \psi_d(\bx, \theta) &= \mE \left[ \nabla_{\bx} \psi(\bx) \right],
    \\
    \label{eq:gradients_Sigma_stein}
    \nabla_{\Sigma_\theta} \psi_d(\bx, \theta) &= \frac{1}{2} \mE \left[ \nabla_{\bx\bx}^2 \psi(\bx) \right].
\end{align}
\end{subequations}
Since $\psi(\bx)$ can be decomposed at the marginal level as
\begin{equation}
\label{eq:collision_factorization}
    \psi(\bx) = \sum_{n=1}^{N-1}  \psi_i(\bx_i) = \sum_{i=1}^{N-1}\left\|\mathbf{h} \left(\bx_{i}\right)\right\|_{\Sigma_{\mathrm{obs}_{i}}}^{2},
\end{equation}
where $\bx_i = M_i \bx$ is a subset of the variables relevant to the $i$th factor as defined in \eqref{eq:mapping_from_joint_to_factor}. Applying \eqref{eq:collision_factorization} to \eqref{eq:gradients_mu_stein}, we obtain
\begin{align}
\label{eq:update_stein_mu}
    \mE_{q}[\nabla_{\bx} \psi(\bx)] = \sum_{i=1}^{N-1} M_{i}^{\mathrm{T}} E_{q_{i}}\left[\nabla_{\bx_i} \psi_{i}\left(\bx_{i}\right)\right]. 
\end{align}
For the $i$-th factor, the gradient and expectation simplify to depend only on $\bx_i$, 
since $\psi_i$ is a function of $\bx_i$ alone. A similar result holds for 
\eqref{eq:gradients_Sigma_stein} 
\begin{align}
    \mE_{q}\left[\nabla_{\bx\bx}^2 \psi(\bx)\right] 
    &= \sum_{i=1}^{N-1} M_{i}^{\mathrm{T}} \mE_{q_{i}}\left[\nabla_{\bx_i\bx_i}^2 \psi_{i}\left(\bx_i\right)\right]M_{i}
    \label{eq:update_stein_Sigma}
\end{align}
Applying Stein's Lemma again to \eqref{eq:update_stein_mu} and \eqref{eq:update_stein_Sigma} gives
\begin{subequations}
    \label{eq:gradients_psid_factorize}
    \begin{align}
        \nabla_{\mu_\theta} \psi_d(\bx, \theta) &= \sum_{i=1}^{N-1} M_{i}^{\mathrm{T}}\nabla_{\mu_{\theta}^{i}} \psi^d_i(\bx_i, \theta_i),
        \label{eq:gradients_psid_mu_factorize}
        \\
        \nabla_{\Sigma_\theta} \psi_d(\bx, \theta) &= \sum_{i=1}^{N-1} M_{i}^{\mathrm{T}} \nabla_{\Sigma_{\theta}^{i}} \psi^d_i(\bx_i, \theta_i) M_{i}.
        \label{eq:gradients_psid_Sigma_factorize}
    \end{align}
\end{subequations}
The factorization of $\psi_d$ allows us to decompose the computation of its gradients across marginal distributions. Combined with parallel processing on a GPU, this structure significantly improves the computational efficiency of the algorithm.

\subsection{Sparse Quadratures for Expectation Estimations}
\label{sec:sparse_GH}
Our primary goal is to estimate the expectation \eqref{eq:gradients_psid_factorize} on the marginal levels. Gauss-Hermite quadrature approximations are widely used in filtering literature~\cite{barfoot2024state}. They are accurate and efficient for lower-dimensional problems. However, a vanilla tensor product of the full-grid sigma points scales \textit{exponentially} with the dimension, making it computationally infeasible for higher-dimensional problems~\cite{gerstner1998numerical}. 

Sparse-grid quadrature rules (Smolyak's rules) \cite{heiss2008likelihood} ignore the cross terms among different dimensions to mitigate the issue, leading to \textit{polynomial} dimensional dependence. This study adopts Smolyak's rules in expectation estimations. To integrate a function $\varphi(\bx)$
\begin{equation*}
    \int \varphi(\bx) \cN(\bx; m, P) d\bx,
    \label{eq:target_integration}
\end{equation*}
G-H quadrature methods consist of the following steps
\begin{enumerate}
    \item Compute the $p$ roots of a $p\; th$ order Hermite polynomial, also denoted as \textit{sigma points}: $\xi = [\xi_1, \dots, \xi_p].$

    \item Compute the \textit{weights}: $W_i = \frac{p!}{p^2[H_{p-1}(\xi_i)]}, i=1, \dots, p.$

    \item Approximate: 
    \begin{equation}
    \label{eq:GH_quadrature_matmul}
        \int \varphi(\bx) \cN(\bx;m,P) d\bx \approx \sum_i W_i \varphi(\sqrt{P}\xi_i + m).
    \end{equation}
\end{enumerate}
    
\begin{lem}[Smolyak's rule complexity]
\label{lem:Smolyak_GH_quadrature}
    For an $n$-variate function, the number of computations needed by a sparse grid quadrature that is exact for $p=2k_q-1$ total polynomial order is bounded by \cite{heiss2008likelihood}
    \[
    \frac{e^{k_q}}{(k_q - 1)!}n^{k_q}
    \]
    which has a polynomial dependence on the desired precision.
\end{lem}

{
\setlength{\algomargin}{1.55em}
\begin{algorithm}[h]
    \caption{P-GVIMP (LTV system).}
    \label{alg:distributed-gvimp}
    \DontPrintSemicolon

    \SetKwInOut{Input}{input}\SetKwInOut{Output}{output}
    \Input{LTV system $\{(A_t, a_t, B_t)\}_{t=1}^{N}$; Number of factors $L$; Upper bound of KL divergence $\epsilon$
    }
    \Output{Optimized trajectory distribution $\cN(\mu_\theta^*, \Sigma_\theta^*)$.}
        
    For $i=0,\dots,N$, compute the state transitional kernel $\Phi_{i,i+1}$, the Grammian $Q_{i,i+1}$, and the prior mean trajectory $\mu_{i}$.

    Compute prior precision $\bK^{-1}$ and prior mean $\bmu$
        
    \For{$k = 1,2,\ldots$}{

        Compute marginals using GBP \eqref{eq:mapping_from_joint_to_factor}, \eqref{eq:belief_lambda}.
        $\left\{q_{\ell} \sim \mathcal{N}\left(\mu_{\theta}^{\ell}, \Sigma_{\theta}^{\ell}\right)\right\}_{\ell=1}^{L} \leftarrow\left(\mu_{\theta}, \Sigma_{\theta}^{-1}\right)$
        
        Parallel Computing the collision factors.
        
        Mapping the gradients back to the joint level
        $\left(\nabla_{\mu_\theta} \psi_d(\bx), \nabla_{\Sigma_\theta} \psi_d(\bx) \right) \leftarrow\left\{\left(\nabla_{\mu_{\theta}^{\ell}} \psi^d_{\ell}(\bx_{\ell}), \nabla_{\Sigma_{\theta}^{\ell}} \psi^d_{\ell}(\bx_{\ell}), M_{\ell}\right)\right\}_{\ell}$

        Select the largest $\beta_k$ using bisection such that: $\mathrm{KL}(q_{\theta_{k+1}} \| q_{\theta_k}) \leq \epsilon, \quad \Sigma_\theta^{-1} \succ 0$
        
        Compute $\mu_{\theta_{k+1}}$ and $\Sigma_{\theta_{k+1}}^{-1}$ using \eqref{eq:mu_Sigma_update}
        $\left(\mu_{\theta_{k+1}}, \Sigma_{\theta_{k+1}}^{-1}\right) \leftarrow \left(\mu_{\theta_k}, \Sigma_{\theta_{k}}^{-1}, \nabla_{\mu_\theta} \psi_d, \nabla_{\Sigma_\theta} \psi_d, \bK^{-1}, \bmu \right)$
    }
    
\end{algorithm}
}

{
\setlength{\algomargin}{1.55em}
\begin{algorithm}[h]
    \caption{i-P-GVIMP (Nonlinear system).}
    \label{alg:i-gvimp}
    \DontPrintSemicolon

    \SetKwInOut{Input}{input}\SetKwInOut{Output}{output}
    \Input{Nonlinear system $\{(f_t, g_t)\}_{t=1}^{N}$, stepsize $\eta$; Nominal trajectory $\{(\bar{x}^0_t, \bar{\Sigma}^0_t)\}$.
    }
    \Output{Optimized trajectory distribution $\cN(\mu_\theta^*, \Sigma_\theta^*)$.}
        
    \For {$i=0,\dots,N$}{
    Linearize the nonlinear system using SLR
    $\left\{ \left( A^i_t, a^i_t, B^i_t \right) \right\} \leftarrow {\rm SLR}(f_t, g_t, \bar{x}^i_t, \bar{\Sigma}_t^i) $

    Perform P-GVIMP (Algorithm \ref{alg:distributed-gvimp}) on the linearized system
    
    $\left\{ \left( \bar{z}^{i+1}_t, \bar{\Sigma}^{i+1}_{t} \right) \right\} \leftarrow$ \rm P-GVIMP$\left( A^i_t, a^i_t, B^i_t \right) $

    }
    
\end{algorithm}
}

\section{Belief Propagation and GPU-Parallel Collision Checking on Factor Graphs}
\label{sec:p-gvimp}
We introduce the two key components that greatly increase the efficiency of the P-GVIMP algorithm for LTV systems. To begin with, we first state our main assumption on the number of discritizations and the state dimensions.

\begin{assumption}
    The discretization dimension $N$ (over $1000$ for dense trajectories) is far greater than the state space dimension $n$ (usually less than $20$ for robotics systems), i.e., $N \gg n$.
\end{assumption}

\subsection{Belief propagation for marginal covariances update}
\label{sec:GBP}
Computing the marginal updates \eqref{eq:update_stein_mu} and \eqref{eq:update_stein_Sigma} requires computing the marginal mean and covariance of the Gaussian distributions $q_i$. A major bottleneck is computing the marginal covariances. A brute force inverse computation of the joint covariance from $\Sigma^{-1}_\theta$ has cubic complexity $O((N\times n)^3)$. 

Leveraging the sparse factor graph, this section introduces the Gaussian Belief Propagation (GBP) \cite{shental2008gaussian, Ortiz2021visualGBP} to compute the marginals over the factor graph. We write the Gaussian distribution $q\sim \mathcal{N}(\mu_\theta, \Sigma_\theta)$ in its canonical form $q(x)\propto \exp(-\frac{1}{2} x^{\top} \Lambda_\theta x+\eta_\theta^{\top} x)$ where $\Lambda_\theta=\Sigma_\theta^{-1}$ denotes the precision matrix, and $\eta_\theta=\Sigma_\theta^{-1} \mu_\theta$ denotes the information vector.

To simplify the calculation, assume a shifted Gaussian distribution $r \sim \mathcal{N}(0, \Sigma_\theta)$, since only the marginal covariance is desired. Then $r(x)$ is written as
\begin{equation}
\label{eq:zero_mean_Gaussian}
    r(x)\propto \exp(-\frac{1}{2} x^{\top} \Lambda_\theta x)
\end{equation}
The precision matrix $\Lambda_\theta$ has the following sparsity pattern
\begin{equation*}
    \Lambda_\theta \! =  \!
    \begin{bmatrix}
    \Lambda_{00} \!\! & \Lambda_{01} \!\! &   &  &  &\\
    \Lambda_{10} \!\! & \Lambda_{11} \!\! & \Lambda_{12} \!\! &  &  &\\
     & \Lambda_{11} \!\! & \Lambda_{12} \!\! &  &  &\\
     &  &  & \dots &  &\\
     &  &  &  &\!\! \Lambda_{(N-1)(N-1)} \!\! & \Lambda_{(N-1)N}\\
     &  &  &  &\!\! \Lambda_{N(N-1)} \!\! & \Lambda_{NN}
    \end{bmatrix},
\end{equation*} 
and the Gaussian distribution \eqref{eq:zero_mean_Gaussian} is factorized into $r(x)\propto \prod_{i=0}^N f_i(x_{i}) \  \prod_{i=0}^{N-1} f_i(x_{i}, x_{i+1})$ with
\begin{equation*}
    \begin{aligned}
    f_i(x_{i})= & \exp \left(-\frac{1}{2} x_i^T \Lambda_{ii} x_i\right), \\
    f_{i,i+1}(x_{i}, x_{i+1})= & \exp \left(-\frac{1}{2} x_{i,i+1}^T \Lambda_{i,i+1} x_{i,i+1}\right),
    \end{aligned}
\end{equation*}
and
\begin{equation}
    x_{i,i+1} = \begin{bmatrix}  x_i \\ x_{i+1}  \end{bmatrix}, \quad \Lambda_{i,i+1} = \begin{bmatrix}  0 & \Lambda_{i(i+1)} \\ \Lambda_{(i+1)i} & 0 \end{bmatrix}
    \nonumber
\end{equation}

Gaussian Belief Propagation is an algorithm for calculating the marginals of a joint distribution via local message passing between nodes in a factor graph \cite{Ortiz2021visualGBP}. Message passing on the factor graph falls into two types. The message passed from variables to factors, denoted as $m_{x_{i} \rightarrow f_{j}}$, and the message passed from factors to variables, denoted as $m_{f_{j} \rightarrow x_{i}}$. Messages are obtained by
\begin{subequations}
\label{eq:messages}
\begin{align}
    m_{x_{i} \rightarrow f_{j}}  & =  \prod_{s \in N(i) \backslash j} m_{f_{s} \rightarrow x_{i}} , \\
    m_{f_{j} \rightarrow x_{i}}  & = \sum_{X_{j} \backslash x_{i}} f_{j}\left(X_{j}\right) \prod_{k \in N(j) \backslash i} m_{x_{k} \rightarrow f_{j}} , 
\end{align}
\end{subequations}
where $N(i)$ denotes all the neighboring factors of $x_i$, $f_{j}\left(X_{j}\right)$ denotes the potential of the factor $f_j$.

After computing all the messages in the factor graph, the beliefs of variables are obtained by taking the product of incoming messages: $b_{i}\left(x_i\right)=\prod_{s \in N(i)} m_{f_{s} \rightarrow x_i}$. Since the messages here are also Gaussian, we obtain the belief parameters $\Lambda_{bi}$ by
\begin{equation}
\label{eq:belief_lambda}
    \Lambda_{b_{i}}=\sum_{s \in N(i)} \Lambda_{f_{s} \rightarrow x_i}
\end{equation}

The time complexity of inverting an $n \times n$ matrix is $O(n^3)$, leading to total complexity $O(N\times n^3)$ for GBP.

\subsection{Parallel Collision Checking on Dense Trajectory}

One of the most computationally heavy modules for motion planning algorithms is the collision-checking module on a dense trajectory \cite{schulman2014motion, mukadam2018continuous}. This is also the case in our framework, where the collision-checking is required in a probabilistic sense, i.e., the goal is to compute the expected collision costs with respect to the proposal distribution.

\begin{figure}[th]
    \centering
    \includegraphics[width=0.9\linewidth]{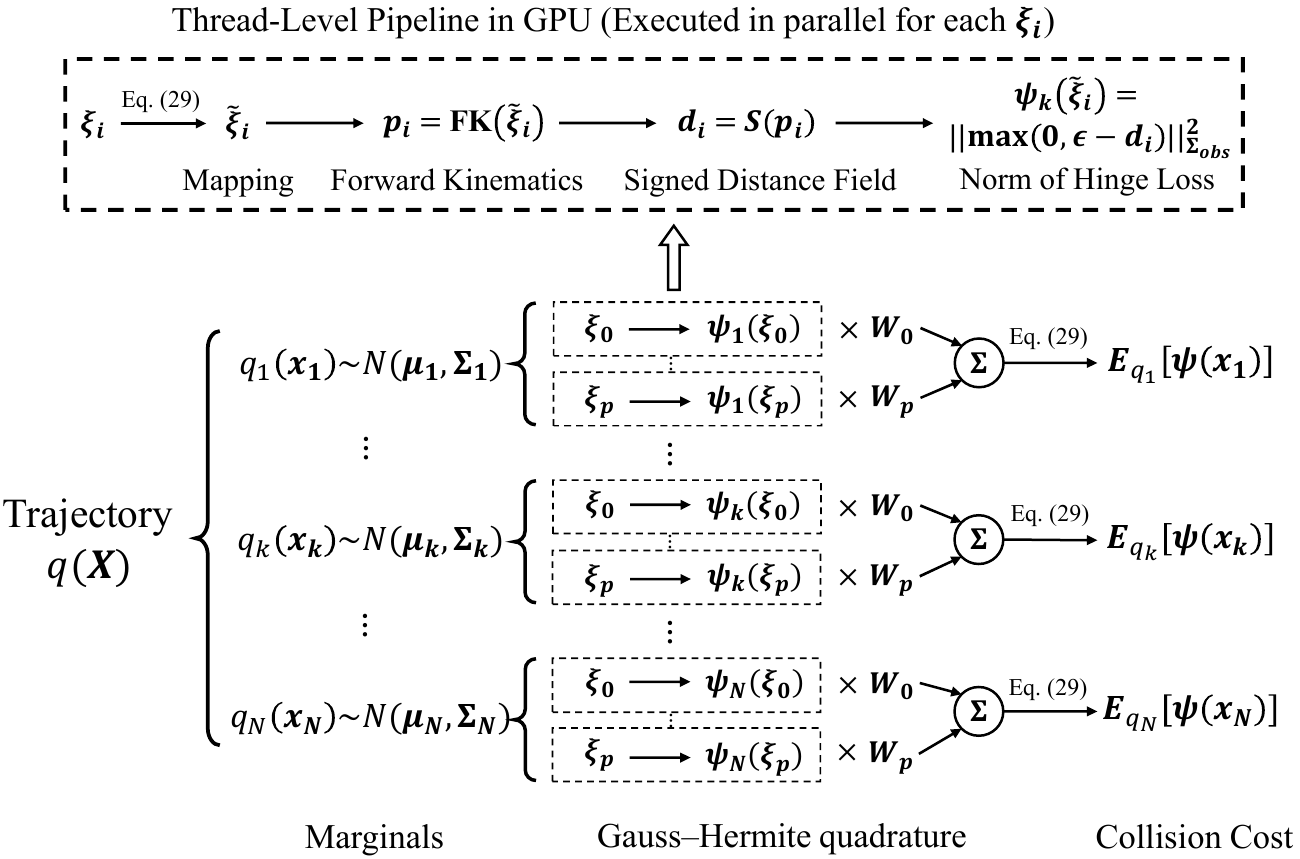}
    \caption{Parallel collision checking}
    \label{fig:parallel_collision_checking}
\end{figure}

We compute the expected collision costs by distributing the computation to marginals in the factor graph (c.f., Section~\ref{sec:factorization}). At marginal levels, we deploy the sparse GH quadratures to compute the marginal collision costs (c.f., Section~\ref{sec:sparse_GH}). The expectation approximation is done by computing \eqref{eq:GH_quadrature_matmul}. Thus, we distribute the expectation approximation by distributing the matrix multiplication \eqref{eq:GH_quadrature_matmul} on the GPU. The graphical demonstration of the parallel collision checking is shown in Figure~\ref{fig:parallel_collision_checking}.

\subsection{The P-GVIMP Algorithm and Complexity Analysis}
The P-GVIMP algorithm is summarized in the Algorithm \ref{alg:distributed-gvimp}. It serves as a sub-routine in planning for nonlinear systems. 

{\em \textbf{Complexity Analysis.}} Denote the state dimension as $n$, and the time discretization number as $N+1$. The factor graph has $L=2N+3$ factors, the maximum dimension being $d=2n$. The main computational burdens are: (1) computing the nonlinear factors; and (2) computing marginal covariances.

Serial computation of the nonlinear factors has linear dependence on $N$, with total complexity $O(N\times n^{k_q})$, where $k_q$ is the desired precision in the GH-quadrature \cite{yu2023stochastic}. The parallel nonlinear factor computation on a GPU reduces the complexity to $O(n^{k_q})$. Experimentally, to obtain a precise estimate of the collision costs, $k_q \geq 6$ suffices. On the other hand, the complexity of computing the marginal covariances using GBP on the tree-structured factor graph is $O(N \times n^3)$, which becomes the new bottleneck. The total algorithm complexity changes from $O(N\times n^{k_q})$ to $O(N\times n^{3})$.

\section{Iterative P-GVIMP for Nonlinear Systems}
\label{sec:nonlinear_control}
Real robot dynamics are nonlinear. The P-GVIMP algorithm described in the previous sections is constrained in LTV systems. This section introduces our proposed Iterative P-GVIMP Algorithm (i-P-GVIMP) to handle nonlinearity. 

\subsection{Gaussian Assumptions for Nonlinear Transition}
The Gaussian assumption is widely used in robotics applications~\cite{barfoot2024state, dellaert2021factor}. It assumes the state distribution follows
\begin{subequations}
\label{eq:Gaussian_assumptions}
    \begin{align}
    \frac{d\bar{X}_t}{d t} &= \mE_{\mathcal{N}}\left[ f(X_t) \right],
    \\
    \frac{d {\rm Cov}(X_t)}{d t} &= \mE_{\mathcal{N}}[ f(X_t) ( X_t - \bar{X}_t )^T ] + \mE_{\mathcal{N}}[ ( X_t - \bar{X}_t ) f(X_t)^T ]
    \nonumber
    \\
    &+ \mE_{\mathcal{N}}\left[ g_t(X_t) g_t(X_t)^T \right],
\end{align}
\end{subequations}
where $\mE_{\mathcal{N}}\left[ \psi(\cdot) \right]$ denotes expectations under Gaussian distributions. We adopt the Gaussian assumptions \eqref{eq:Gaussian_assumptions} and propose an iterative statistical linearization of the nonlinear system around a nominal trajectory. The P-GVIMP Algorithm \ref{alg:distributed-gvimp} is then performed on the obtained linearized LTV system. The results are used in the next iteration for the linearization. This process is performed until convergence.

\subsection{Iterative Statistical Linear Regression (SLR).} 
Statistical linear regression (SLR) \cite{arasaratnam2007discrete} is a statistical method to approximate the nonlinear stochastic processes locally. The linearization results in an LTV system depending on the state at which the linearization is performed. 

Assuming the closed-loop nonlinear stochastic dynamics at time $t=t_i$ as $X_{i+1} = f_{\rm cl}(X_i)$, then SLR finds a linear model $\hat{X}_{i+1} = A_i X_i + a_i$ for $(A_i, a_i)$ to minimize the statistical error
\begin{equation}
\label{eq:SLR_objective}
    \{A_i, a_i\} = \arg\min \mE \left[ \lVert X_{i+1} - \hat{X}_{i+1} \rVert^2 \right].
\end{equation}
Taking the first-order derivative of \eqref{eq:SLR_objective} with respect to $a_i$ and letting the derivative be zero, we arrive at
\begin{align}
\label{eq:SLR_best_ai}
    a_i^* = \mE\left[ f_{\rm cl}\left( X_i \right) \right] - A_i^* \mE\left[ X_i \right].
\end{align}
substituting \eqref{eq:SLR_best_ai} into \eqref{eq:SLR_objective}, the gradients over $A_i$ is
\begin{equation*}
    \left( f_{\rm cl}(X_i) - \mE\left[ f_{\rm cl}(X_i) \right] - A_i \left( X_i - \mE\left[ X_i \right] \right)  \right) \left( X_i - \mE\left[ X_i \right] \right)^T.
\end{equation*}
Letting the above be zero leads to 
\begin{equation}
\label{eq:SLR_best_Ai}
    A_i^* = P_{yx} P_{xx}^{-1}, 
\end{equation}
where $P_{yx}$, $P_{xx}$ are defined respectively as
\begin{align}
    P_{yx} &\triangleq  \left( f_{\rm cl}(X_i) - \mE\left[ f_{\rm cl}(X_i) \right] \right) \left( X_i - \mE\left[ X_i \right] \right)^T
    \\
    P_{xx} &\triangleq \left(X_i - \mE\left[ X_i \right]\right) \left(X_i - \mE\left[ X_i \right]\right)^T.
\end{align}
In \eqref{eq:SLR_best_ai} and \eqref{eq:SLR_best_Ai}, we use the sparse GH-quadrature rules in Lemma \ref{lem:Smolyak_GH_quadrature} to compute the expectations of a nonlinear function. At time $t_i$, for a given approximation precision, $k_q$, we compute the weights and sigma points $\{ (W^i_l, \xi^i_l) \}_{l=1}^{n^{k_q}}$, and perform the quadrature approximation in \eqref{eq:SLR_best_ai} and \eqref{eq:SLR_best_Ai}
\[
\mE\left[ X_{i+1} \right] = \mE\left[ f_{\rm cl}(X_i) \right] \approx \sum_{l=1}^{n^{k_q}} W_l f(\sqrt{P_{xx}} \; \xi_l + \mE\left[ X_i \right]).
\]
This procedure is iteratively performed forward in time. The iterative P-GVIMP is summarized in Algorithm \ref{alg:i-gvimp}. Our proposed paradigm has profound connections with the previous works \cite{yu2023gaussian, yu2023stochastic}, where we targeted to solve the efficiency issues and for nonlinear systems. Table~\ref{tab:comparisons} summarizes the connections and differences between P-GVIMP and GVIMP.    

\begin{table}[h]
    \centering
    \begingroup
    \begin{tabular}{|c|c|c|}
        \hline
         & GVIMP & P-GVIMP \\ \hline
        Paradigm & \multicolumn{2}{c|}{Planning-as-inference} \\ \hline
        Structure & \multicolumn{2}{c|}{Sparse Planning Factor Graph} \\ \hline
        System              & Linear                   &    Nonlinear \\ \hline
        Algorithm           & NGD    & SLR + KL-Proximal \\ \hline
        Complexity     & $O(N \times n^{k_q})$             &  $O(N\times n^{3})$   \\ \hline
        Hardware            & CPU                               &  GPU            \\ \hline
        \end{tabular}
    \caption{Comparison between the proposed paradigm and the previous works on GVIMP \cite{yu2023gaussian, yu2023stochastic}.}
    \label{tab:comparisons}
    \endgroup
\end{table}

\section{Numerical Experiments}
\label{sec:experiments}
This section presents the numerical experiment results for the proposed method. The experiments are aimed at understanding:
\begin{itemize}[leftmargin=*]
    \item[] \textbf{Q1:} Efficiency gains by parallel computation on GPUs versus serial computation on CPUs;
    \item[] \textbf{Q2:} The effectiveness of the KL-proximal algorithm in high-DOF robot motion planning tasks compared GPMP2;
    \item[] \textbf{Q3:} The effectiveness of the iterative P-GVIMP algorithm in motion planning tasks for nonlinear dynamical systems.
\end{itemize}
This section answers these questions sequentially and concludes with ablation and comparison studies.

\subsection{Efficiency Improvement through Parallel Computation}
The first experiment aims to demonstrate the improved computational efficiency achieved by parallel collision-checking on a GPU and marginal covariance computation using GBP. 

{\em (a) Parallel collision-checking.}
To demonstrate the improvement in collision-checking efficiency on dense trajectories, we compare the computation time of expected collision factors between the proposed GPU-parallel 
implementation and a serial baseline \cite{yu2023stochastic}. The experiments involve 2D and 3D 
point robots, as well as a 7-DOF WAM arm robot, modeled as a linear time-invariant (LTI) system
\begin{equation*}
    A_t = \begin{bmatrix}
    0 & I \\
    0 & 0
    \end{bmatrix},~ 
    a_t=\begin{bmatrix}
        0\\
        0
    \end{bmatrix},~ 
    B_t=\begin{bmatrix}
    0 \\ I
    \end{bmatrix}.
\end{equation*}

\begin{table}[h]
\centering
\resizebox{\columnwidth}{!}{
\begin{tabular}{|c|c|c|c|}
\hline
 & $2$D Point Robot & $3$D Point Robot & $7$-DoF WAM\\ \hline
Serial   & $85.40$ \text{ms} & $479.7$ \text{ms} & $8010.5$ \text{ms} \\ \hline
Parallel & $14.80$ \text{ms} & $37.60$ \text{ms} & $306.16$ \text{ms}   \\ \hline
Improvement & $\textbf{82.67 \%}$  &  $\textbf{92.16 \%}$ & $\textbf{96.18 \%}$ \\ \hline
\end{tabular}
}
\caption{Implementation time comparison for Cost Expectation Estimations (quadrature precision degree: $k_q=10$ for 2D and 3D Point Robots, and $k_q=6$ for 7-DoF WAM Arm; time discretization $N = 750$). }
\label{tab:cost_time_comparison}
\end{table}

\begin{figure} [ht]
\centering
    \includegraphics[width=0.9\linewidth]{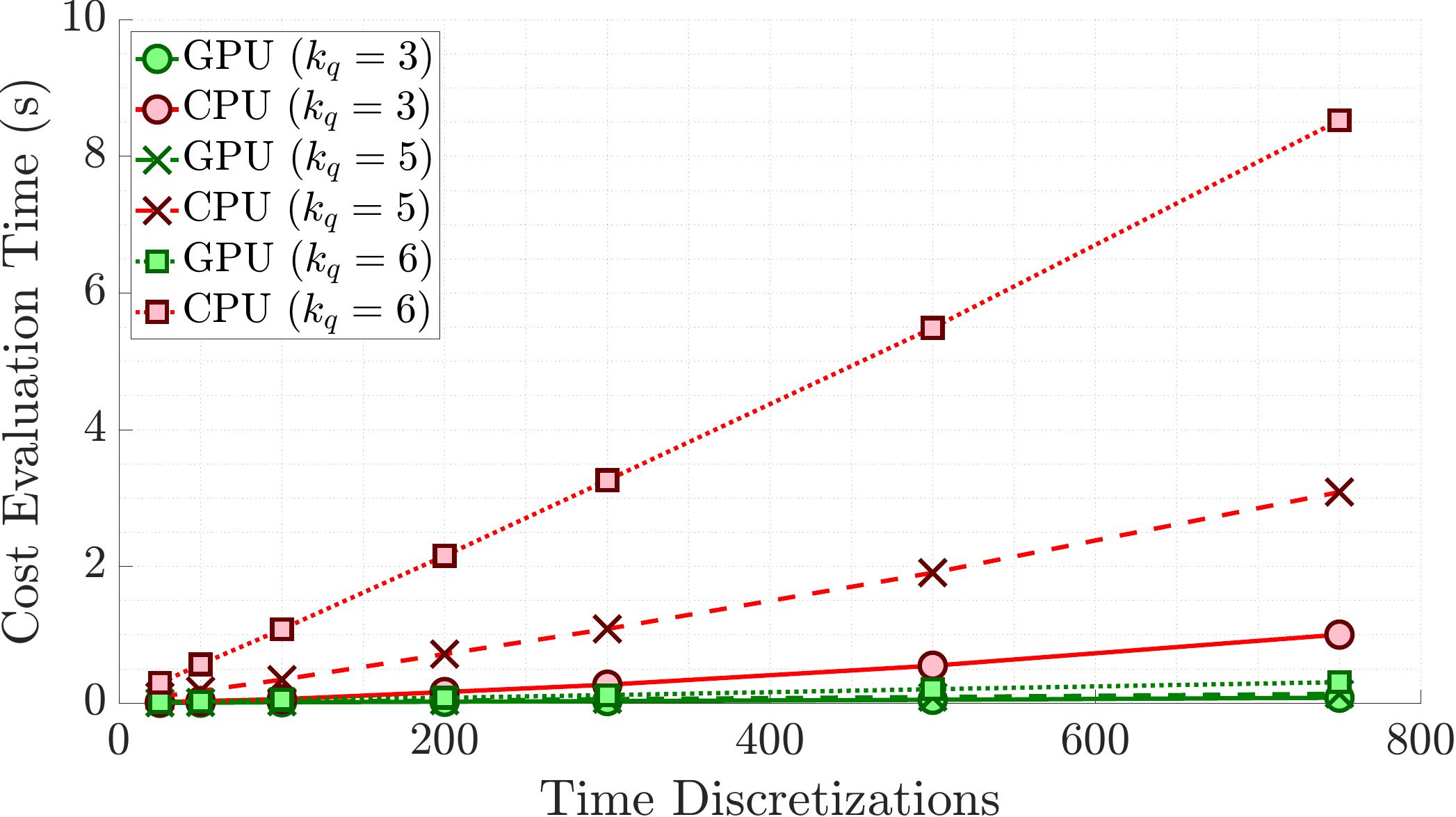}
    \caption{$7$-DOF WAM expected collision cost computation.}
  \label{fig:time_curve_collision}
\end{figure}

The results appear in Table~\ref{tab:cost_time_comparison}. On a dense trajectory with time discretization $N=750$, the GPU-parallel implementation achieves over $95\%$ reduction in computation time compared with the CPU-based serial implementation. Figure~\ref{fig:time_curve_collision} reports the computation time for expected collision-checking costs under different time discretizations. The parallel implementation exhibits substantially lower sensitivity to the discretization level.

\begin{figure}[th]
\centering
    \includegraphics[width=0.9\linewidth]{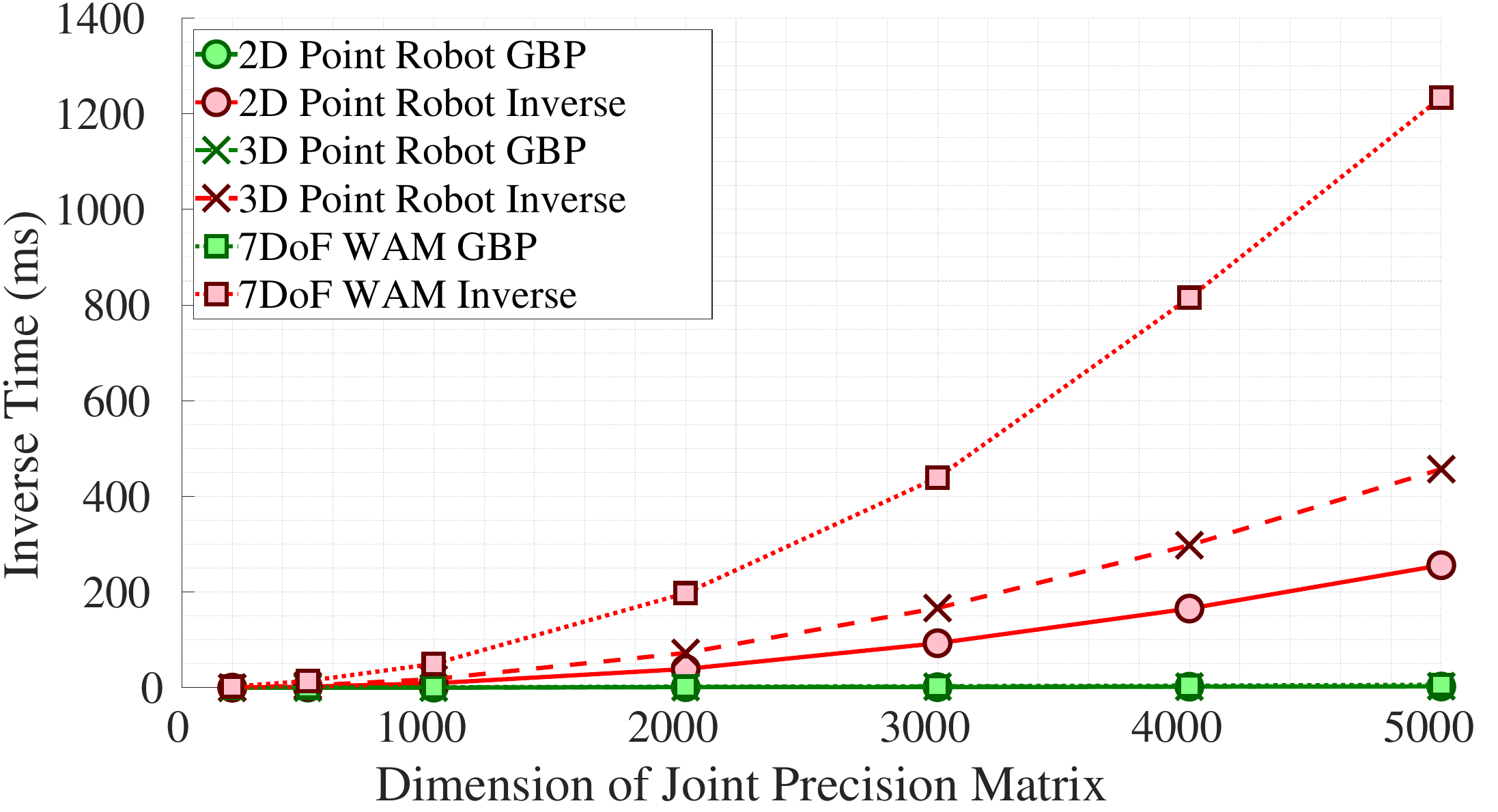}
    \caption{Time-Dimension curve for precision matrix inversion. }
    \label{fig:Inverse_time_curve}
\end{figure}

{\em (b) GBP for computing the marginal covariances.}
Next, we demonstrate the efficiency gains in computing marginal covariances using GBP. Our proposed GBP approach is compared against the brute-force approach of directly inverting the precision matrix $\Sigma_{\theta}^{-1}$. The comparison results appear in Table~\ref{tab:Inverse_comparison} and Figure~\ref{fig:Inverse_time_curve}. GBP exhibits linear dependence on time discretizations $N$, whereas direct inversion scales cubically ($N^3$). For a 7-DOF WAM, GBP achieves a $99.5\%$ improvement in efficiency.

\begin{table}[h]
\centering
\resizebox{\columnwidth}{!}{
\begin{tabular}{|c|c|c|c|}
\hline
 & $2$D Point Robot & $3$D Point Robot & $7$-DoF WAM\\ \hline
 Brute force & $256.4$ \text{ms} & $456.86$ \text{ms}  & $1234.04$ \text{ms} \\ \hline
GBP & $2.36$ \text{ms} & $3.01$ \text{ms} & $6.04$ \text{ms} \\ \hline
Improvement & $\textbf{99.07\%}$ & $\textbf{99.34\%}$ & \textbf{99.51\%} \\ \hline
\end{tabular}
}
\caption{Implementation Time Comparison for Precision Matrix Inversion (Dimension of Precision: $5000$, $4998$, $4998$). 
}
\label{tab:Inverse_comparison}
\end{table}

{\em (c) The overall optimization time.}
Finally, the overall efficiency of the proposed paradigm is evaluated by recording the total 
optimization time and comparing it against a CPU-based serial implementation 
\cite{yu2023gaussian, yu2023stochastic}. The results appear in Table~\ref{tab:total_cost_time_comparison}. For the 7-DOF planning problems, the proposed approach achieves over $97\%$ improvement in efficiency.

\begin{table}[h]
\centering
\resizebox{\columnwidth}{!}{
\begin{tabular}{|c|c|c|c|}
\hline
 & $2$D Point Robot & $3$D Point Robot & $7$-DoF WAM \\ \hline
Serial   &  $7.38$ s  & $27.07$ s &  $425.03$ s \\ \hline
Parallel &  $0.48$ s  & $1.12$ s  &  $12.25$ s  \\ \hline
Improvement & $\textbf{93.50 \%}$  &  $\textbf{95.86 \%}$ &  $\textbf{97.11 \%}$
\\\hline
\end{tabular}
}
\caption{Implementation Time Comparison for the whole optimization process (quadrature precision degree: $k_q=10$ for 2D and 3D Point Robots, and $k_q=6$ for 7-DoF WAM Arm; time discretization $N = 750$).}
\label{tab:total_cost_time_comparison}
\end{table}

\subsection{Planning Results for the 7-DOF Robot Arms}
\label{sec:arm_experiment}
\begin{figure*}[ht]
\centering
    \begin{subfigure}[ht]{0.23\linewidth}
    \centering
    \includegraphics[width=0.95\linewidth]{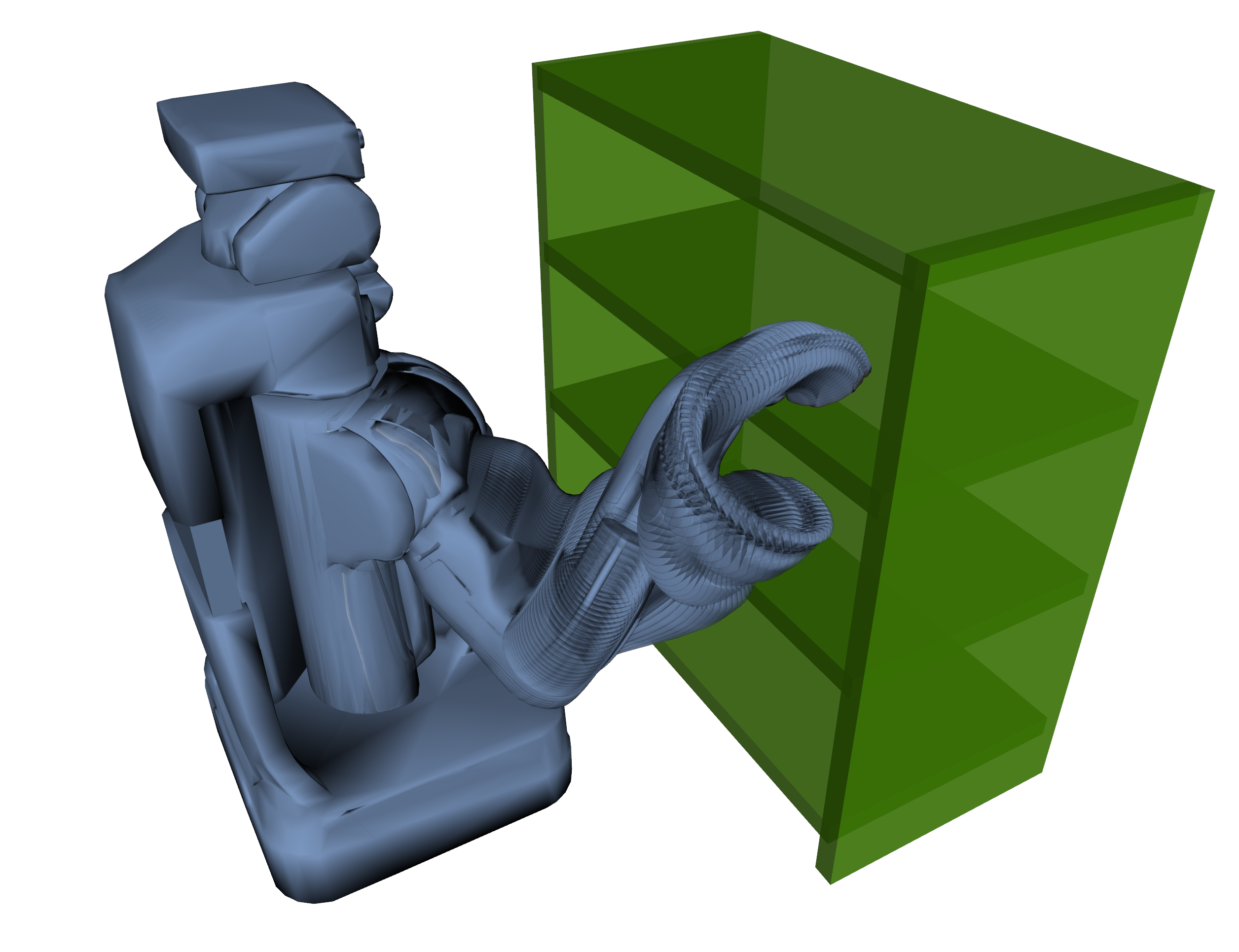}
    \end{subfigure}
    \begin{subfigure}[ht]{0.23\linewidth}
    \centering
    \includegraphics[width=0.95\linewidth]{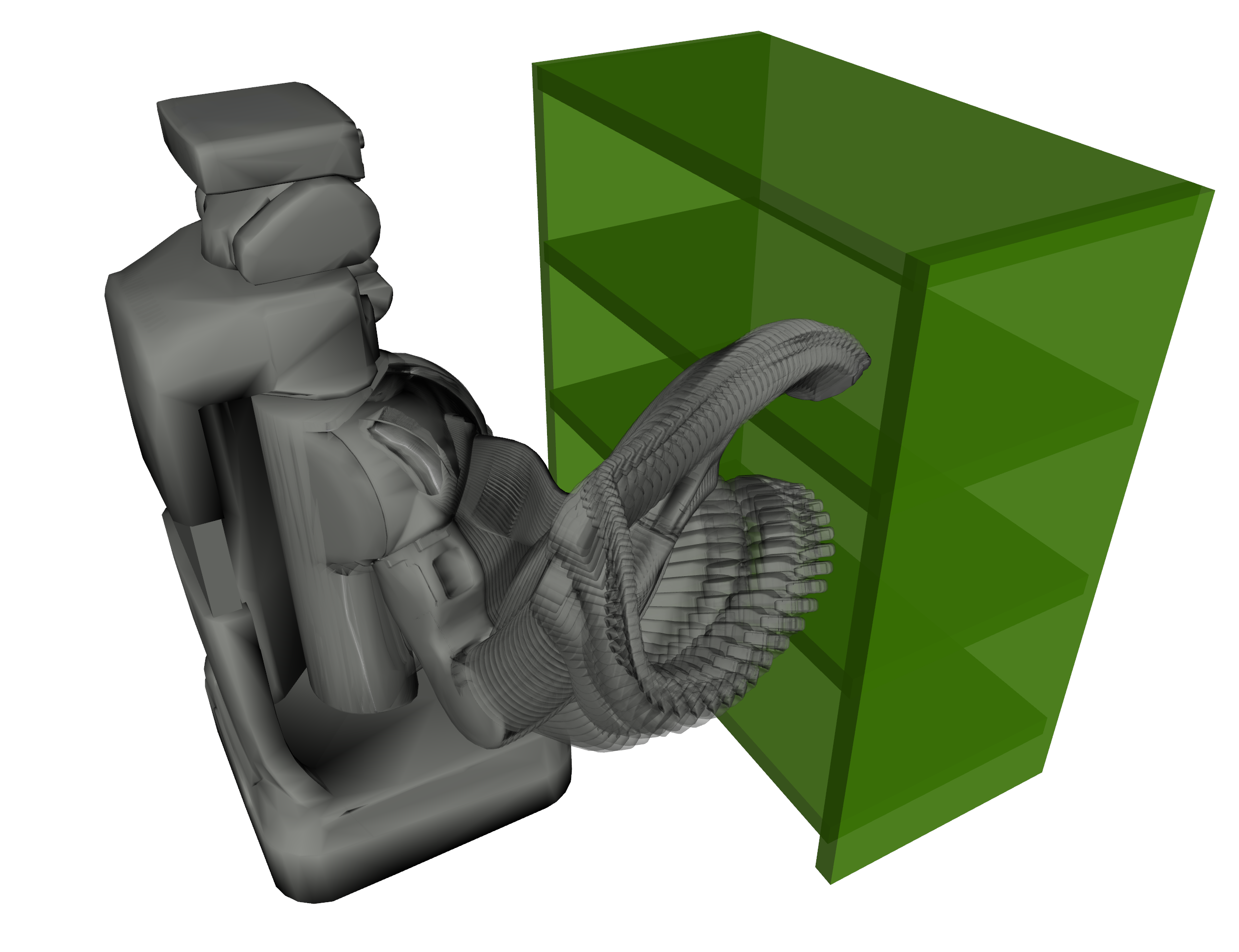}
    \end{subfigure}
    \begin{subfigure}[ht]{0.23\linewidth}
    \centering
    \includegraphics[width=0.95\linewidth]{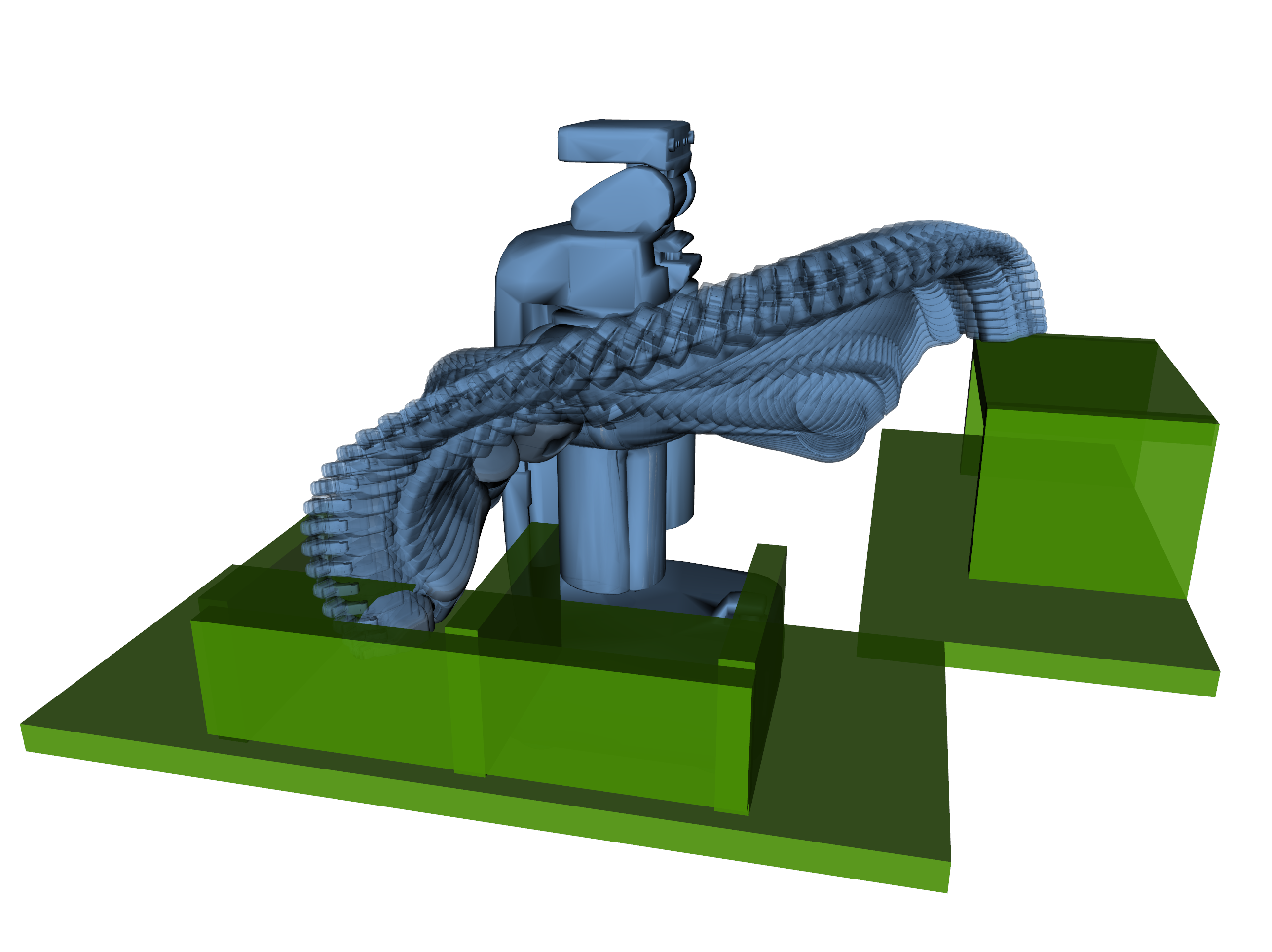}
    \end{subfigure}
    \begin{subfigure}[ht]{0.23\linewidth}
    \centering
    \includegraphics[width=0.95\linewidth]{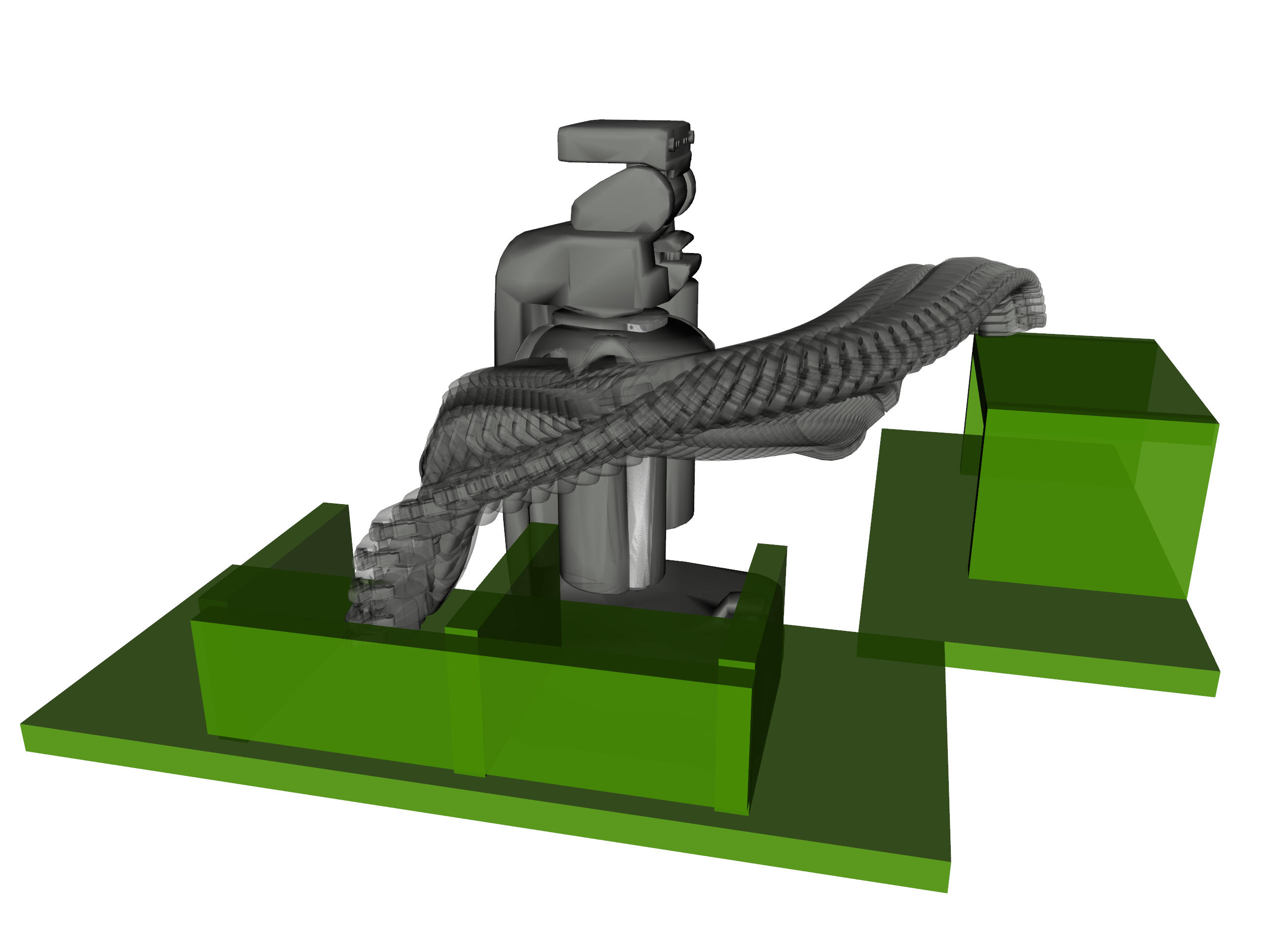}
    \end{subfigure}
  \caption{ Planning results for the PR2 Robot for two tasks with $N=750$ states. The results are obtained from two algorithms: GPMP2 (blue) and KL-Proximal (Gray). For KL-Proximal, the settings are $T = 2.25$ and $2.0$, $\Sigma_{obs} = 7.5I$ and $5I$, respectively, the ratio between temperatures is $\hat{T}_{high}/\hat{T}_{low}=10$, and $r+\epsilon_{sdf}=0.21$}
  \label{fig:pr2_ros}
\end{figure*}

\begin{figure*}[ht]
\centering
    \begin{subfigure}[ht]{0.24\linewidth}
    \centering
    \includegraphics[width=0.95\linewidth]{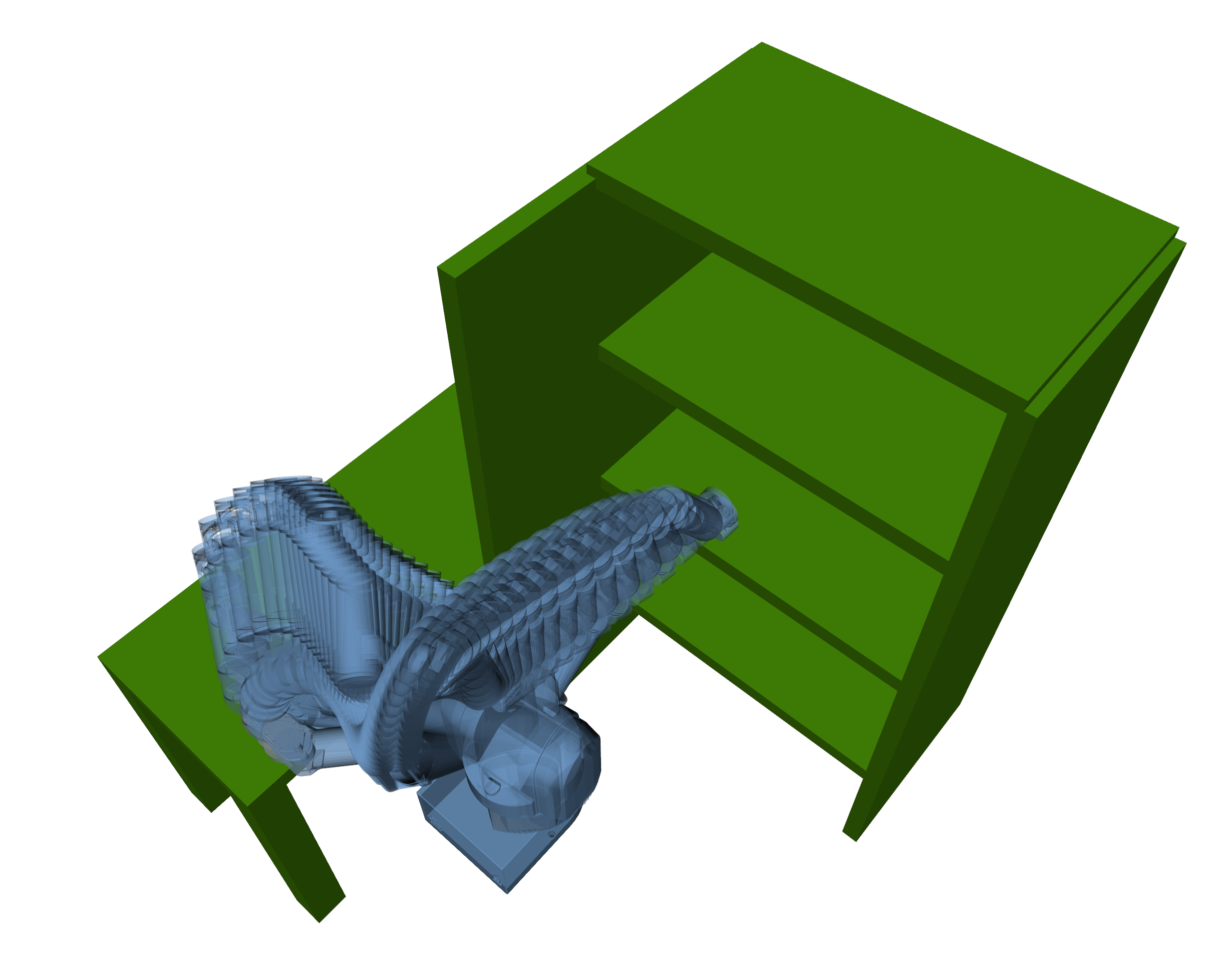}
    \end{subfigure}
    \begin{subfigure}[ht]{0.24\linewidth}
    \centering
    \includegraphics[width=0.95\linewidth]{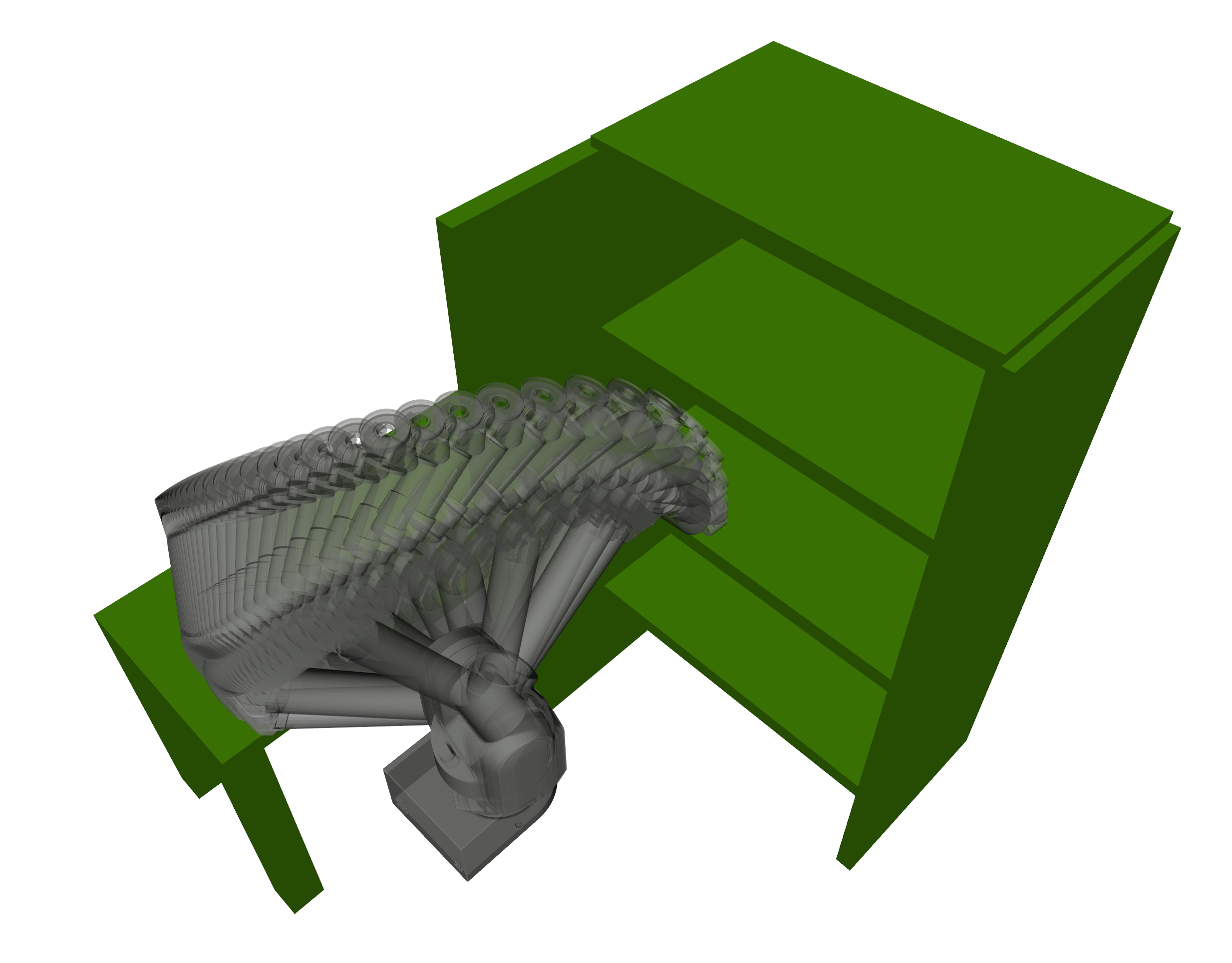}
    \end{subfigure}
    \begin{subfigure}[ht]{0.23\linewidth}
    \centering
    \includegraphics[width=0.95\linewidth]{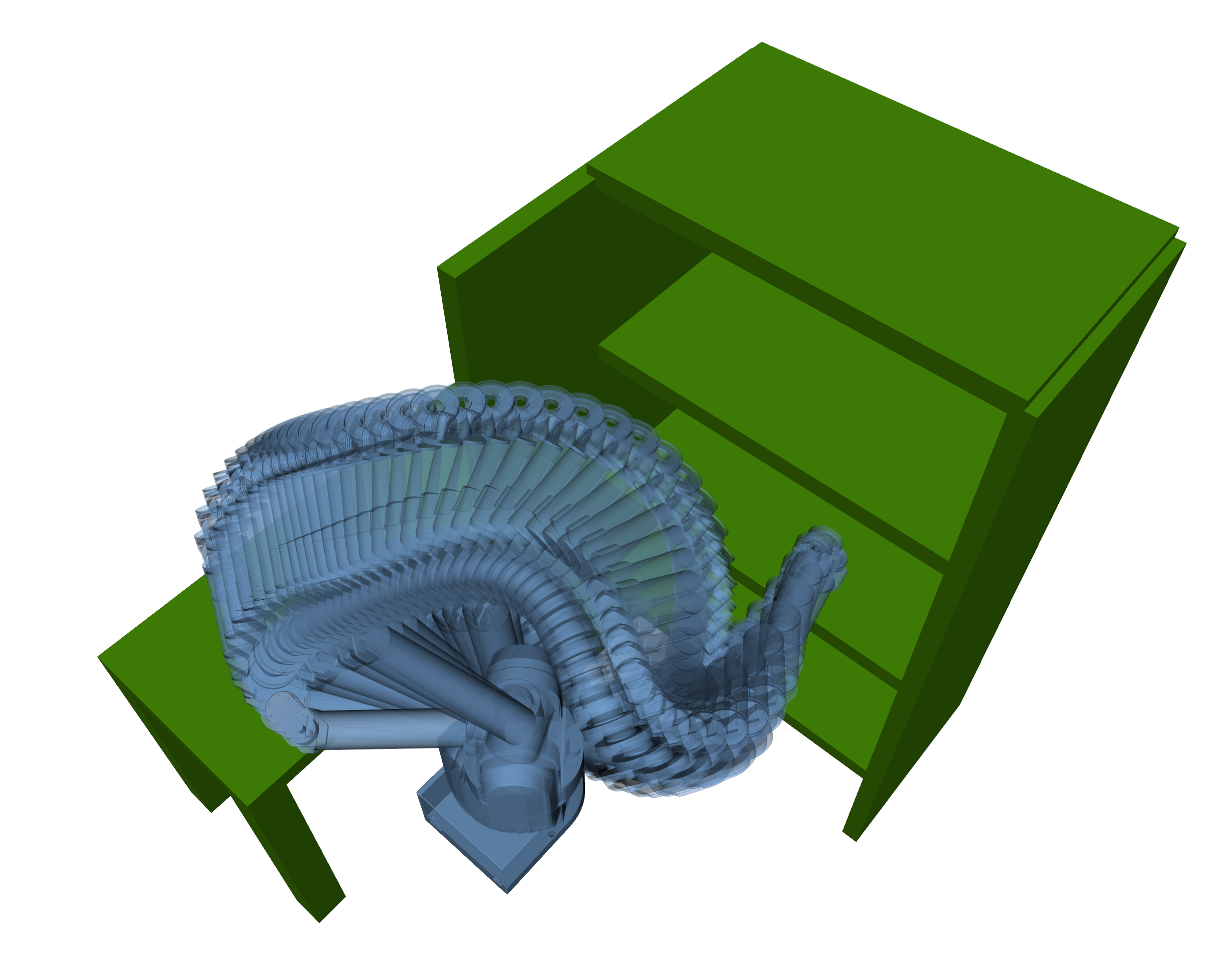}
    \end{subfigure}
    \begin{subfigure}[ht]{0.23\linewidth}
    \centering
    \includegraphics[width=0.95\linewidth]{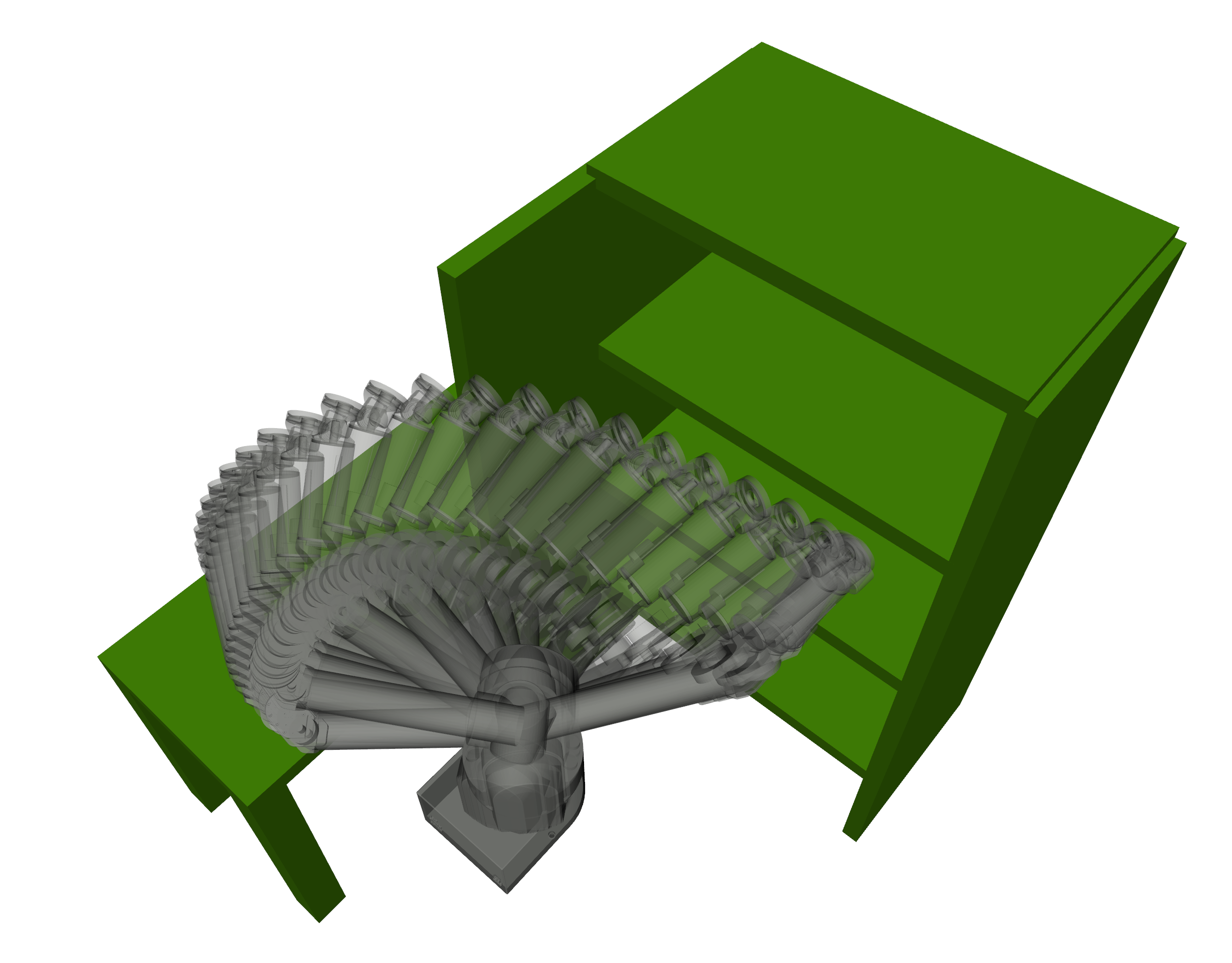}
    \end{subfigure}
  \caption{ Planning results for the WAM Robot for two tasks with $N=750$ states. The results are obtained from two algorithms: GPMP2 (blue) and KL-Proximal (Gray). For KL-Proximal, we set $T = 2.0$ and $2.75$, $\Sigma_{obs} = 20I$ and $19I$, The ratio between temperatures are $\hat{T}_{high}/\hat{T}_{low}=10$ and $15$, respectively, and $r+\epsilon_{sdf}=0.21$}
  \label{fig:wam_ros}
\end{figure*}

Next, the planning results for the WAM robot arm \cite{rooks2006harmonious} and the right arm of the PR2 robot \cite{garage2012pr2} are shown using the Robot Operating System (ROS) \cite{quigley2009ros} and the \textit{Moveit} \cite{coleman2014reducing} planning package. Figure~\ref{fig:wam_ros} shows the planning results for the WAM robot arm, and Figure~\ref{fig:pr2_ros} shows the planning results for the PR2 robot arm. 

\begin{figure*}[ht]
\centering
    \begin{subfigure}[ht]{0.23\linewidth}
    \centering
    \includegraphics[width=0.95\linewidth]{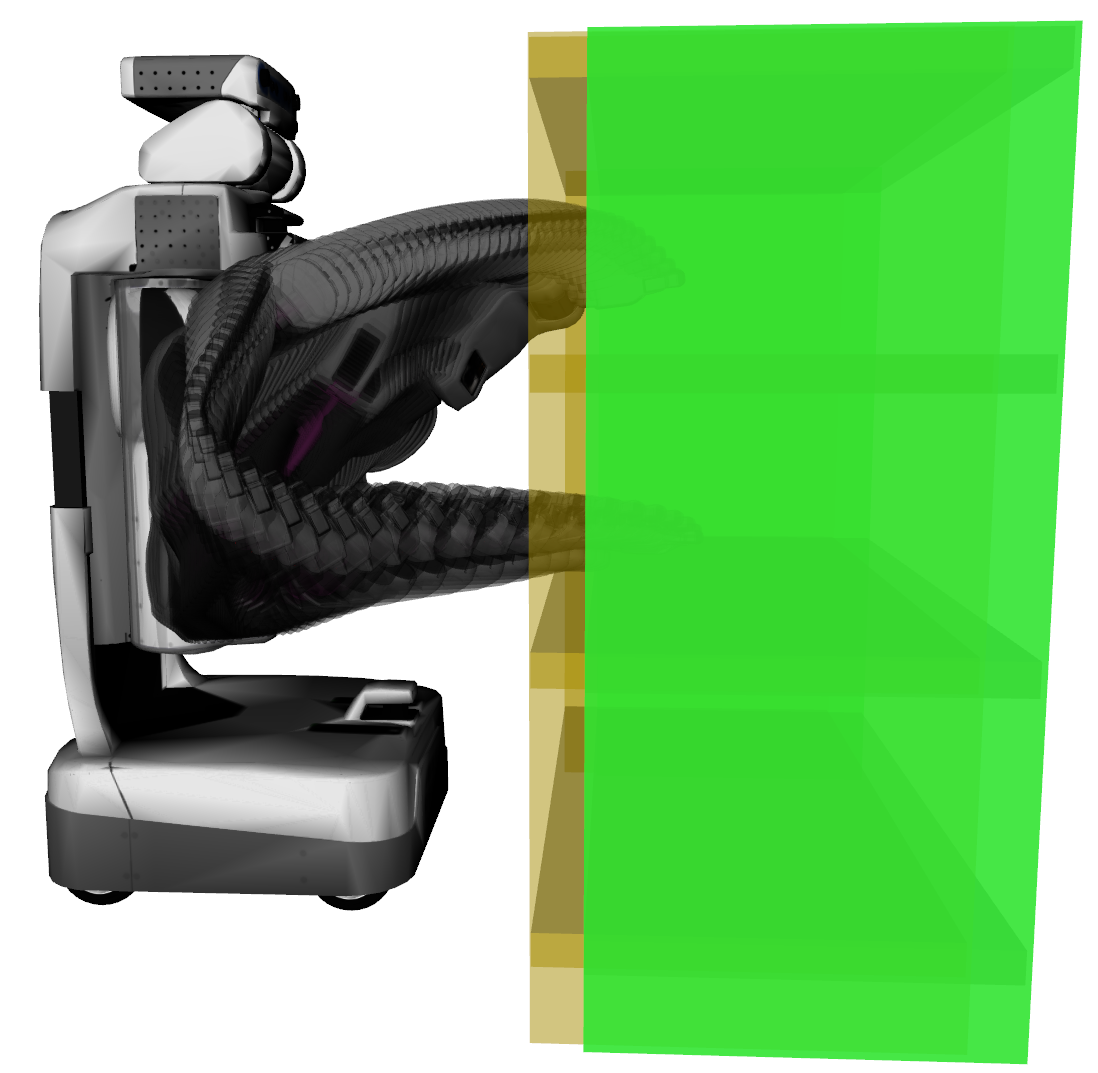}
    \caption{P-GVIMP}
    \end{subfigure}
    \begin{subfigure}[ht]{0.23\linewidth}
    \centering
    \includegraphics[width=0.95\linewidth]{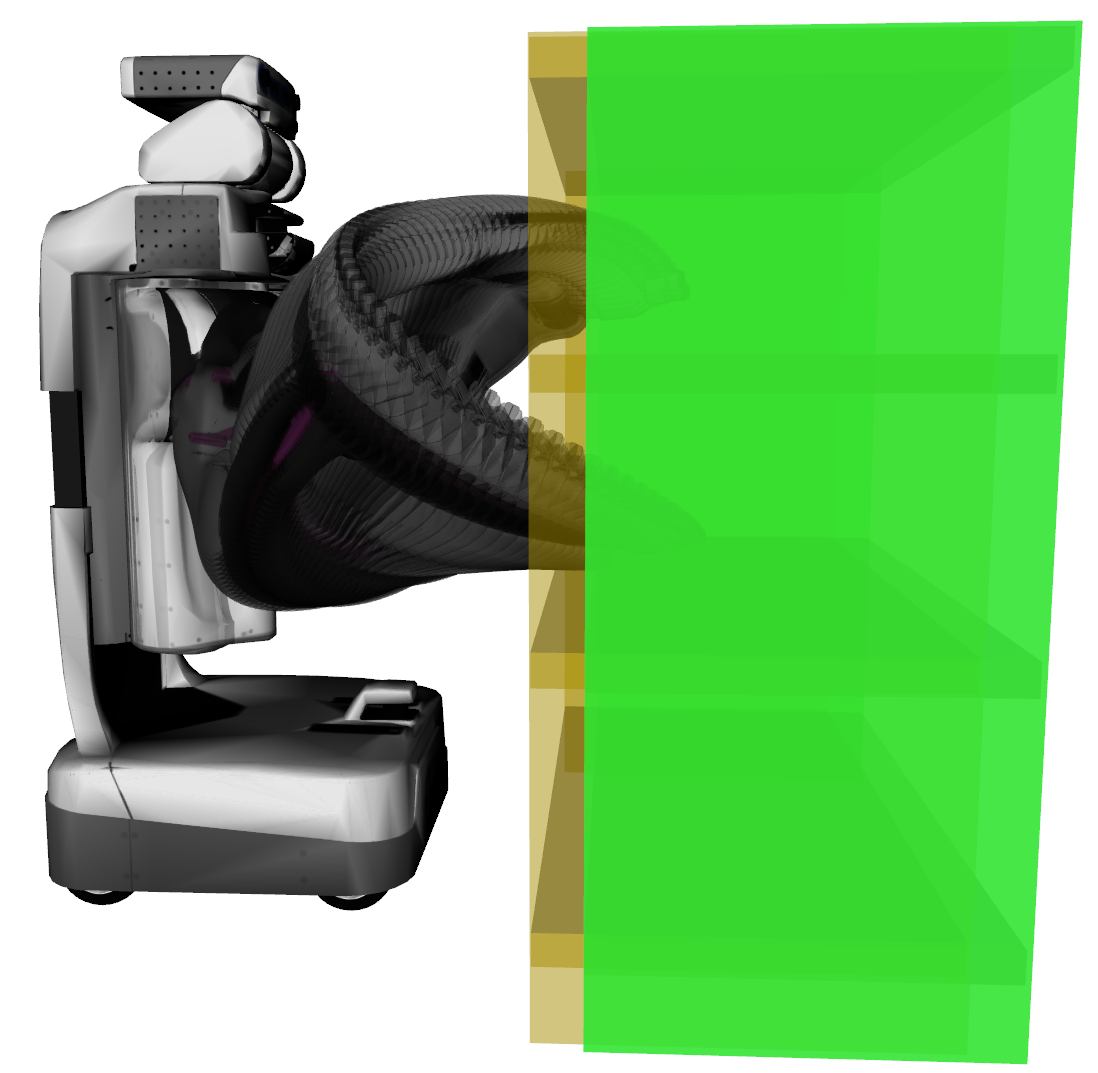}
    \caption{GPMP2}
    \end{subfigure}
    \begin{subfigure}[ht]{0.23\linewidth}
    \centering
    \includegraphics[width=0.95\linewidth]{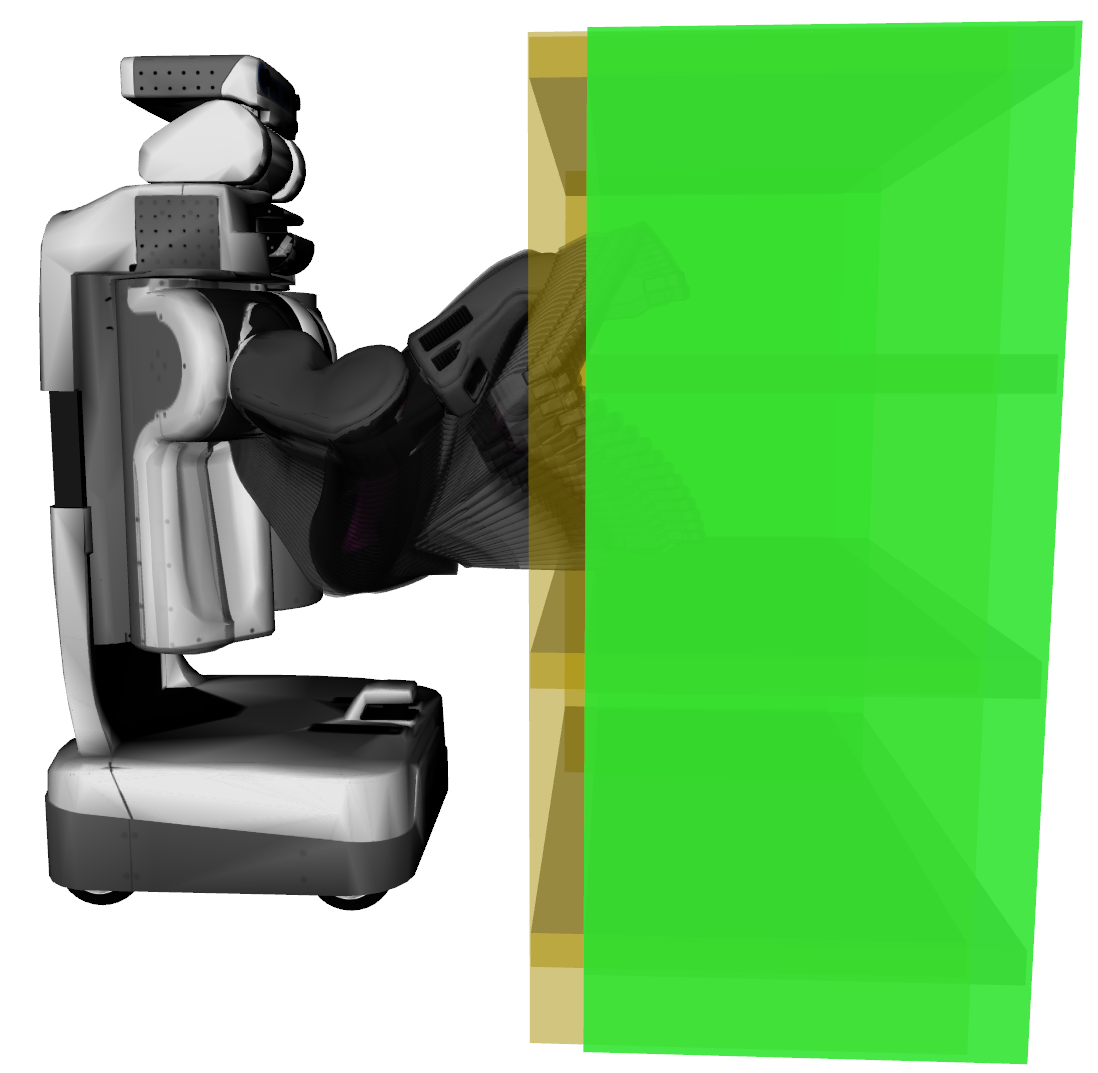}
    \caption{BIT*}
    \end{subfigure}
    \begin{subfigure}[ht]{0.23\linewidth}
    \centering
    \includegraphics[width=0.95\linewidth]{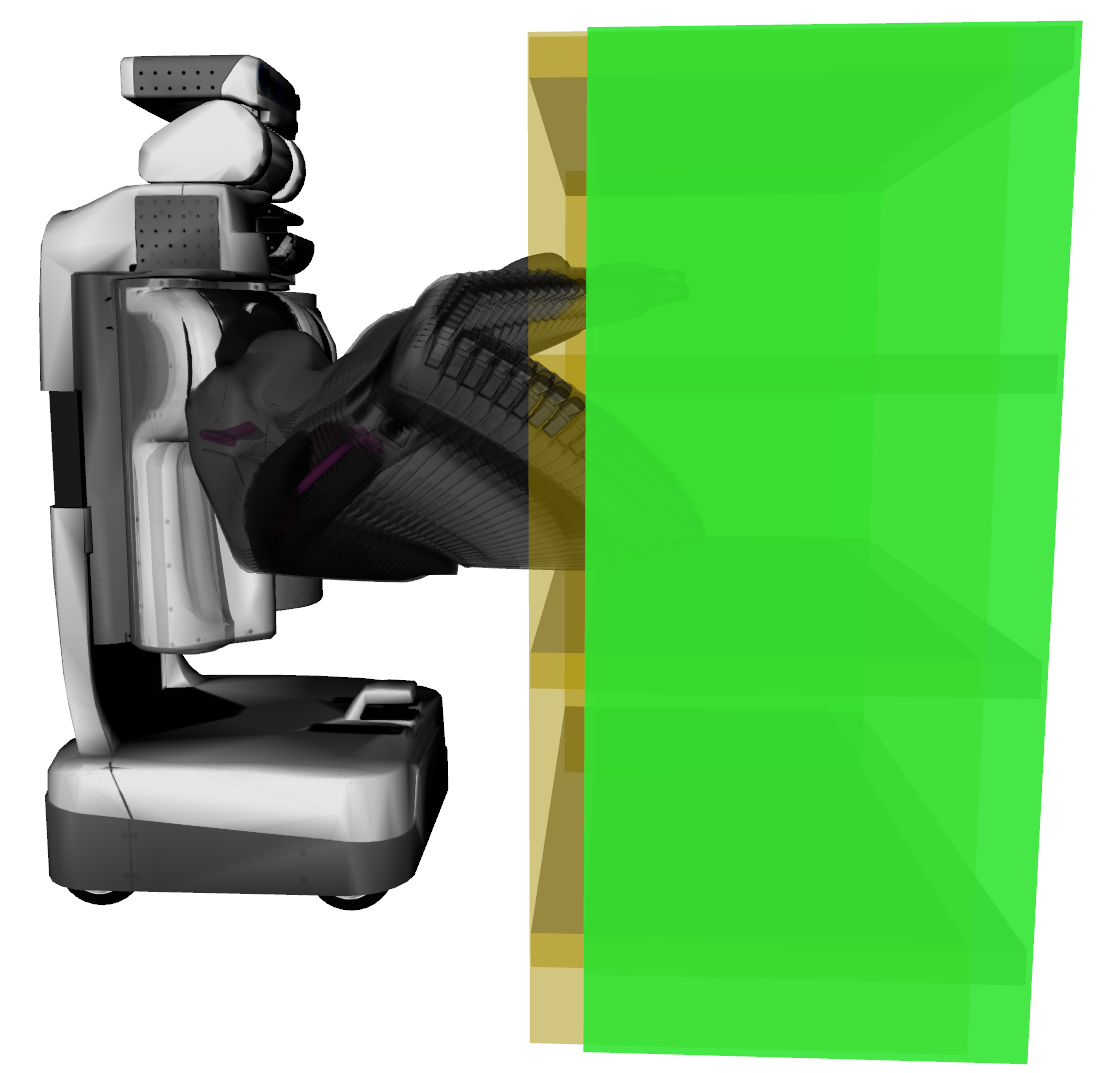}
    \caption{FMT*}
    \end{subfigure}
  \caption{PR2 planning results in an environment with perturbed obstacles. Each subfigure shows the resulting trajectory computed by the corresponding planner: (a) P-GVIMP, (b) GPMP2, (c) BIT*, and (d) FMT. Green obstacles represent the nominal environment, while yellow obstacles indicate the perturbed environment.}
  \label{fig:pr2_disturb}
\end{figure*}

The performance of P-GVIMP under uncertainty is further evaluated by comparing it with GPMP2 \cite{mukadam2018continuous}, BIT* \cite{gammell2015batch}, and FMT* \cite{janson2015fast}. To simulate perception noise, motion plans are first computed in a nominal obstacle environment. The obstacle positions are then perturbed by adding Gaussian noise, after which the plans are executed and the minimum distances from the trajectories to the obstacles are calculated. Figure~\ref{fig:pr2_disturb} illustrates the trajectories generated by all four planners in the same perturbed environment, while Table~\ref{tab:avg_dist} reports the average minimum distance from obstacles, computed over 50 randomly perturbed obstacle configurations.

\begin{table}[t]
\centering
\resizebox{\columnwidth}{!}{
\begin{tabular}{|c|c|c|c|c|}
\hline
 Planner &  P-GVIMP & GPMP2 & BIT* & FMT* 
 \\ 
 \hline
 Average Distance  &  $\textbf{0.0216}$ & $0.0052$ & $-0.0324$ & $-0.0313$\\ \hline
\end{tabular}
}
\caption{Average minimum distance from obstacles for different planners over 50 randomly perturbed environments. 
}
\label{tab:avg_dist}
\end{table}

\subsection{Iterative P-GVIMP for nonlinear dynamical system}
Our proposed algorithm is validated on a planar quadrotor  
\begin{equation}
    \dot{X_t} = \begin{bmatrix}
        v_x \cos(\phi) - v_z \sin(\phi)\\
        v_x \sin(\phi) + v_z\cos(\phi) \\
        \dot\phi \\
        v_z \dot\phi - g\sin(\phi)\\
        -v_x\dot\phi - g\cos(\phi)\\
        0
    \end{bmatrix} + 
    \begin{bmatrix}
        0 & 0 \\
        0 & 0 \\
        0 & 0 \\
        0 & 0 \\
        1/m & 1/m \\
        l/J_q & -l/J_q 
    \end{bmatrix}
    \begin{bmatrix}
        u_1 \\
        u_2
    \end{bmatrix},
    \label{eq:planar_quad}
\end{equation}
where $g$ is the gravity, $m$ represents the mass of the planar quadrotor, $l$ is the length of the body, and $J_q$ is the moment of inertia. $u_1$ and $u_2$ are the two thrust inputs to the system. In all experiments, $m = 1/\sqrt{2}$, $l = \sqrt{2}$, and $J_q = 1$. The system in \eqref{eq:planar_quad} is linearized around a nominal trajectory, and the P-GVIMP algorithm is applied to the resulting LTV system.

\begin{table}[t]
\centering
\begin{tabular}{|c|c|c|c|c|}
\hline
 Iteration &  $1$ & $3$ & $10$ & $20$ 
 \\ 
 \hline
 Norm difference  &  $286.40$ & $16.61$ & $4.33$ & $2.96$\\ \hline
\end{tabular}
\caption{Norm difference between two consecutive linearizations of the nonlinear dynamic. 
}
\label{tab:norm_trj_difference_nolinear}
\end{table}

\begin{figure*}[th]
\centering
    \begin{subfigure}[ht]{0.15\textwidth}
    \centering
    \includegraphics[width=\linewidth]{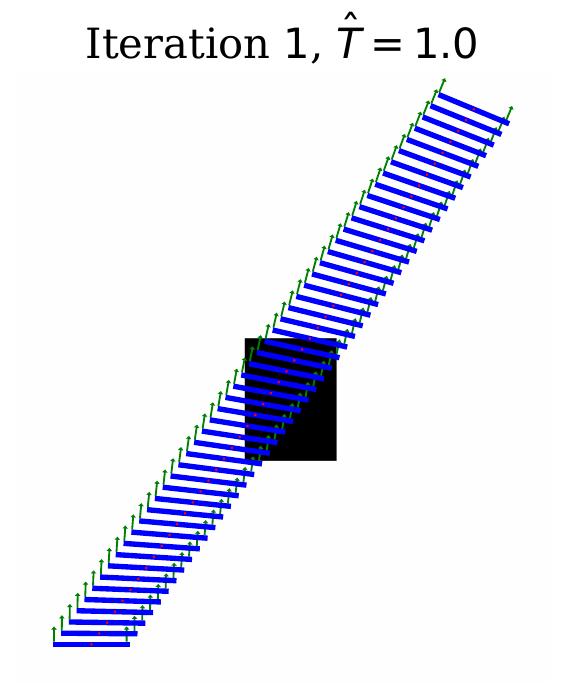}
    \end{subfigure}
    \begin{subfigure}[ht]{0.15\textwidth}
    \centering
    \includegraphics[width=\linewidth]{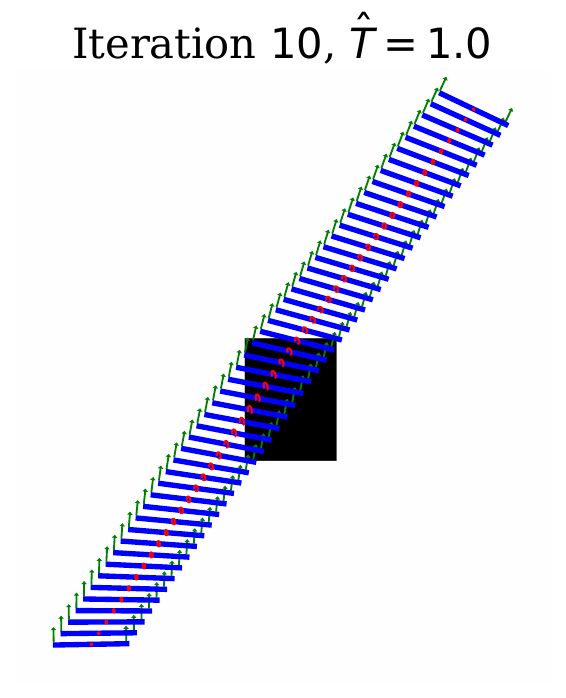}
    \end{subfigure}
    \begin{subfigure}[ht]{0.15\textwidth}
    \centering
    \includegraphics[width=\linewidth]{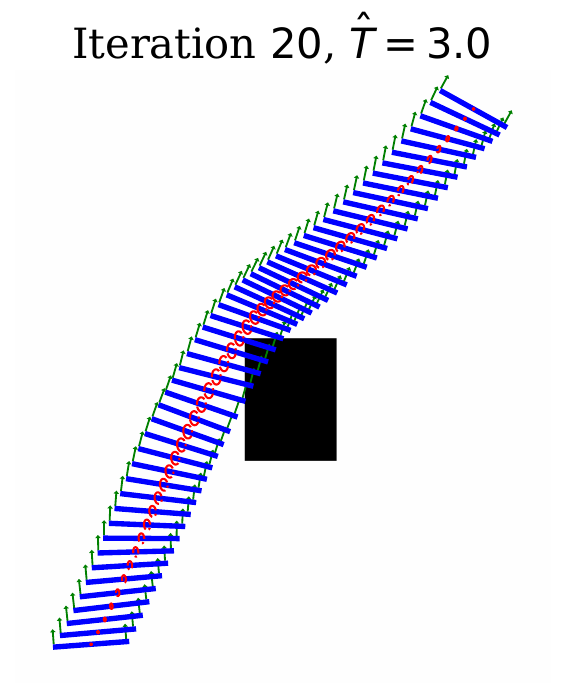}
    \end{subfigure}
    \begin{subfigure}[ht]{0.15\textwidth}
    \centering
    \includegraphics[width=\linewidth]{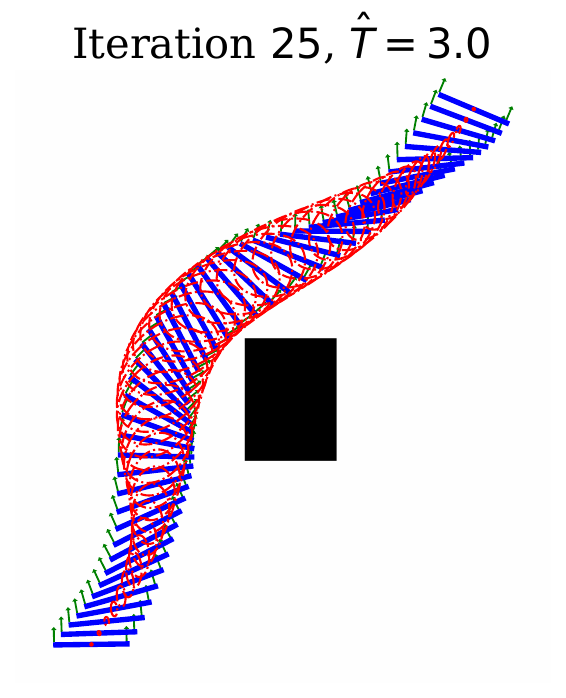}
    \end{subfigure}
  \caption{Convergence of P-GVIMP inner-loop iterations for the linearized planar quadrotor system. }
  \label{fig:ltv_trajectories}
\end{figure*}

\begin{figure*}[th]
\centering
    \begin{subfigure}[ht]{0.15\textwidth}
    \centering
    \includegraphics[width=\linewidth]{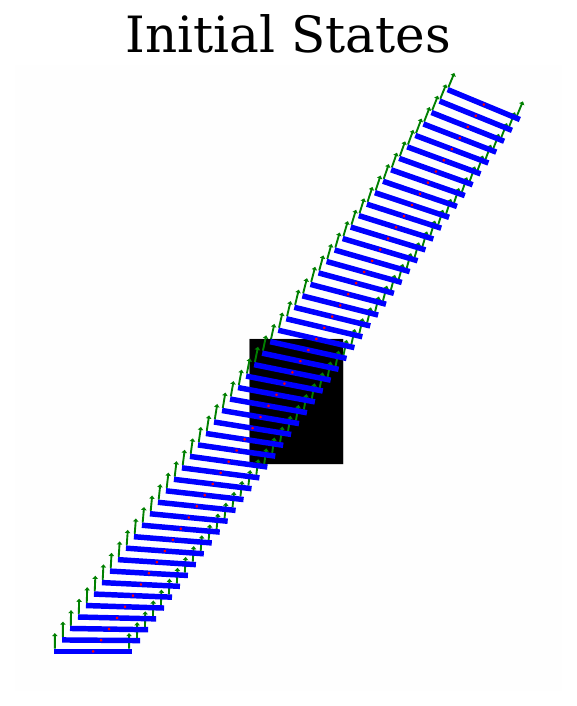}
    \end{subfigure}
    \begin{subfigure}[ht]{0.15\textwidth}
    \centering
    \includegraphics[width=\linewidth]{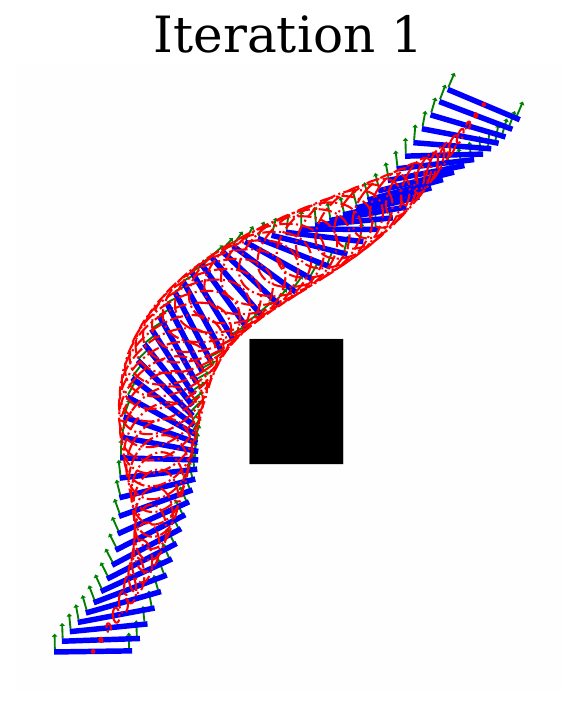}
    \end{subfigure}
    \begin{subfigure}[ht]{0.15\textwidth}
    \centering
    \includegraphics[width=\linewidth]{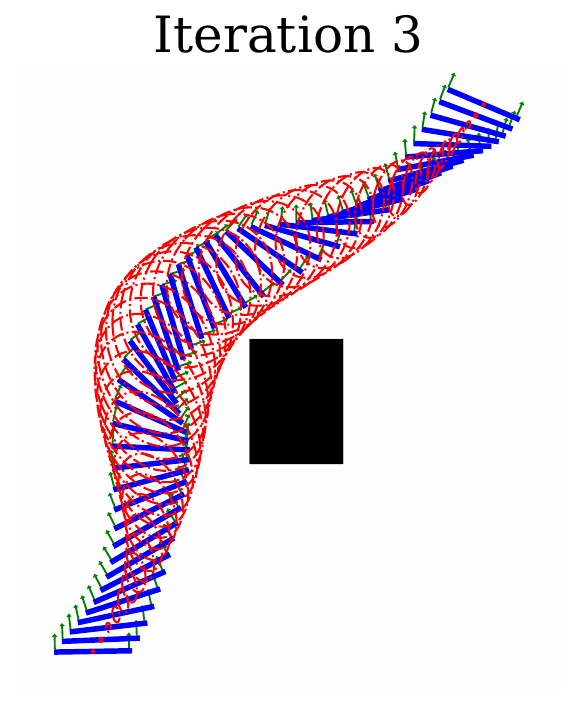}
    \end{subfigure}
    \begin{subfigure}[ht]{0.15\textwidth}
    \centering
    \includegraphics[width=\linewidth]{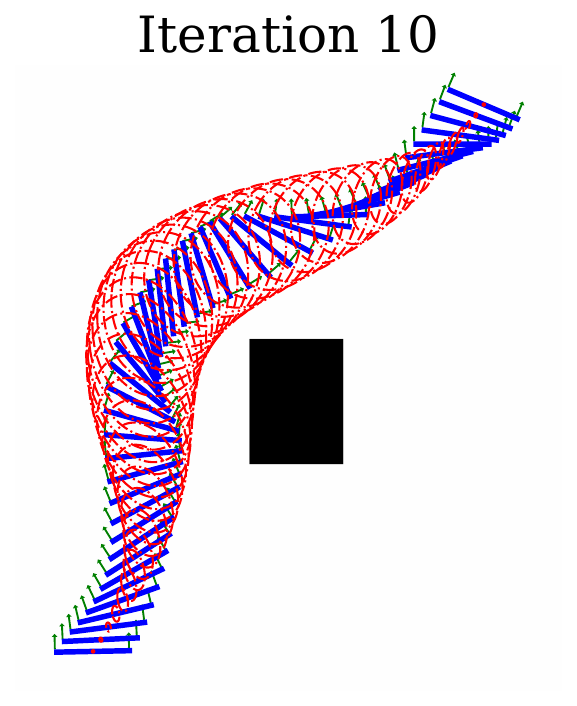}
    \end{subfigure}
    \begin{subfigure}[ht]{0.15\textwidth}
    \centering
    \includegraphics[width=\linewidth]{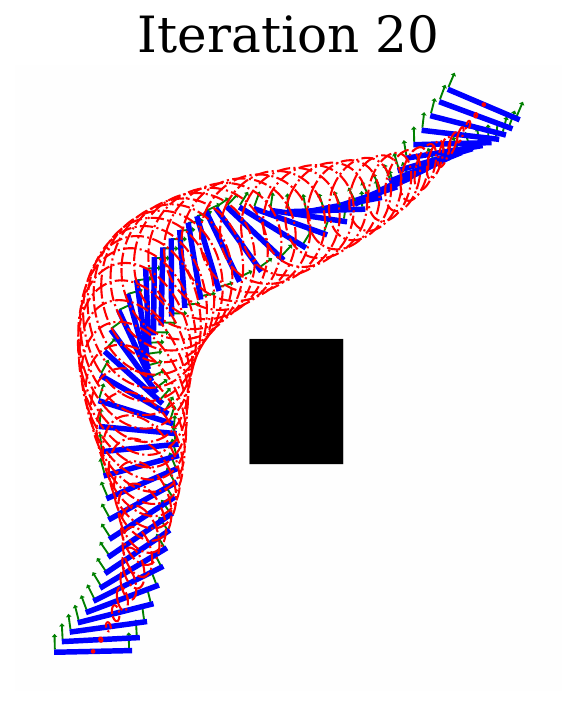}
    \end{subfigure}
  \caption{Convergence of the SLR iterations for the nonlinear planar quadrotor dynamical system. }
  \label{fig:nonlinear_trajectories}
\end{figure*}

\begin{figure}[th]
    \centering
    \includegraphics[width=0.95\linewidth]{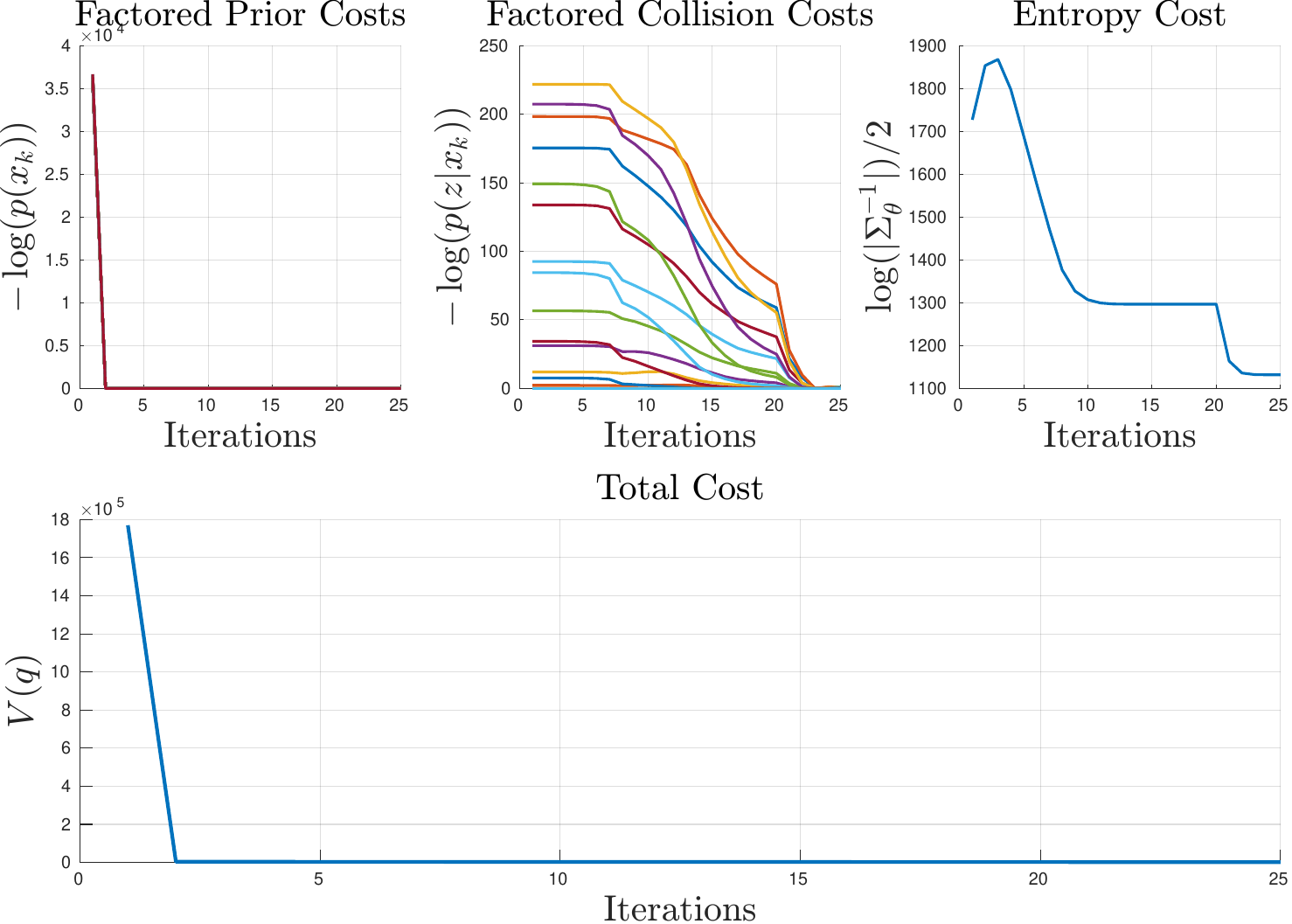}
    \caption{Factorized and total costs in Figure~\ref{fig:ltv_trajectories}.}
    \label{fig:cost_graph}
\end{figure}

{\em (a) Empirical convergence for LTV system.}
We first present results for a \textbf{\textit{linearized}} planar quadrotor system using 
Algorithm~\ref{alg:distributed-gvimp}. Convergence is illustrated in Figure~\ref{fig:ltv_trajectories}, which shows a down-sampled plot of the intermediate solution trajectory distributions with both low and high temperatures $\hat{T}$. The corresponding cost evolutions—including prior costs, collision-factorized costs, and total costs—are reported in Figure~\ref{fig:cost_graph}. After obtaining a collision-free trajectory under low temperatures, the algorithm transitions to a high-temperature phase to emphasize entropy costs. The total cost decreases monotonically throughout both phases.

{\em (b) Planning results for the nonlinear planar quadrotor.}
Next, consider the \textbf{\textit{full}} nonlinear planar quadrotor system \eqref{eq:planar_quad} using the iterative-P-GVIMP planning, i.e., Algorithm \ref{alg:i-gvimp}. The norm of the trajectory difference between consecutive iterations is computed, and the iterations are stopped once the norm difference is below a threshold. The results are recorded in Table~\ref{tab:norm_trj_difference_nolinear}, and the planning results are shown in Figure~\ref{fig:nonlinear_trajectories}.

\begin{figure*}[th]
\centering
    \begin{subfigure}[ht]{0.26\linewidth}
    \centering
    \includegraphics[width=\linewidth]{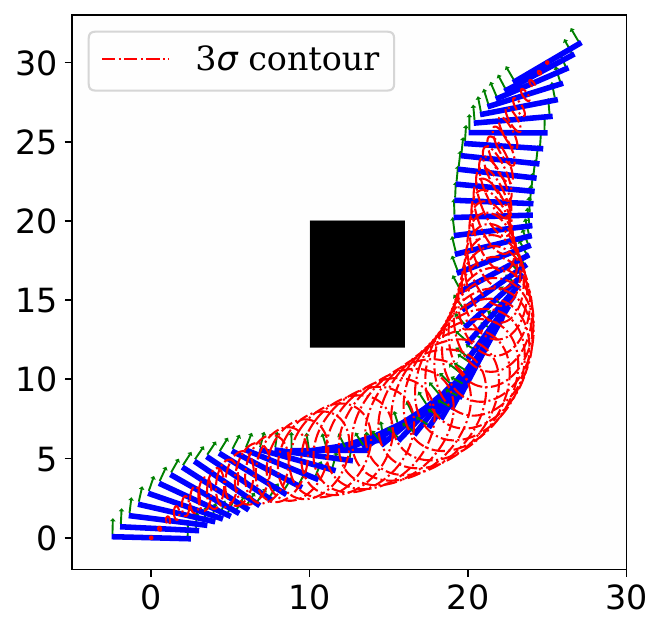}
    \caption{Experiment $1$.}
    \end{subfigure}
    \hfill
    \begin{subfigure}[ht]{0.26\linewidth}
    \centering
    \includegraphics[width=\linewidth]{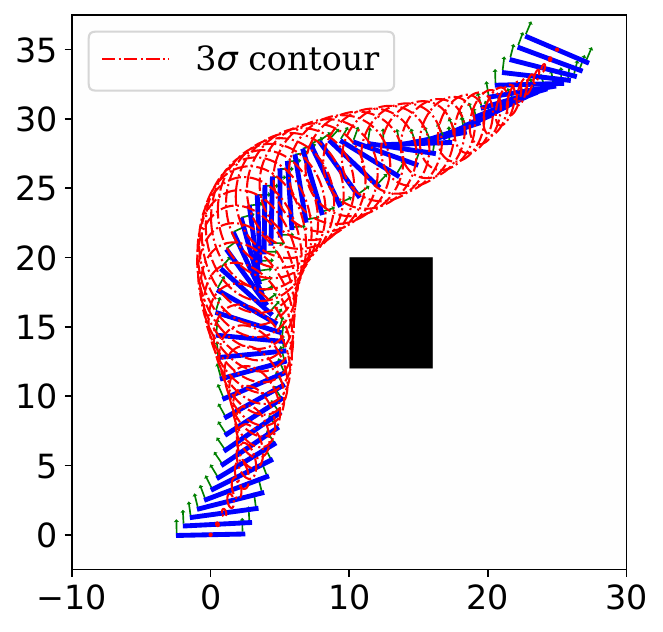}
    \caption{Experiment $2$.}
    \end{subfigure}
    \hfill
    \begin{subfigure}[ht]{0.21\linewidth}
    \centering
    \includegraphics[width=\linewidth]{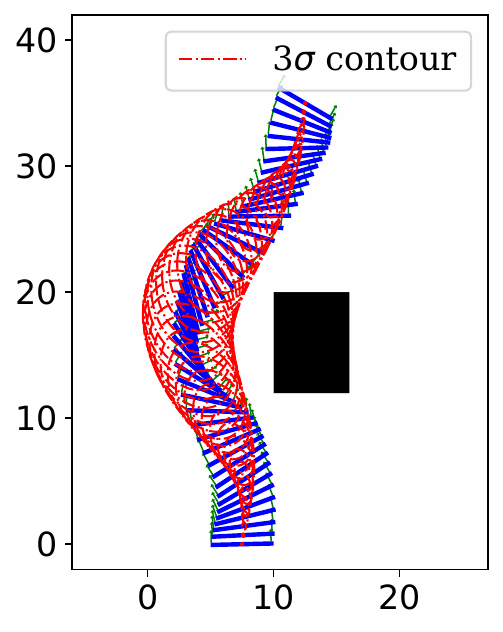}
    \caption{Experiment $3$.}
    \end{subfigure}
    \hfill
    \begin{subfigure}[ht]{0.21\linewidth}
    \centering
    \includegraphics[width=\linewidth]{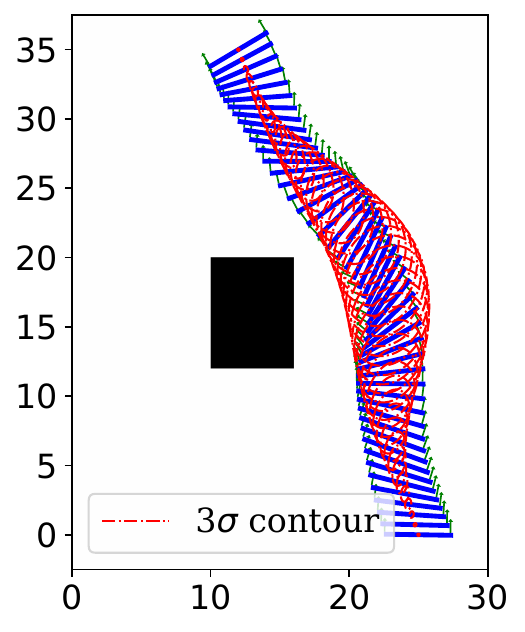}
    \caption{Experiment $4$.}
    \end{subfigure}
    \hfill
  \caption{Results of the i-P-GVIMP for the nonlinear planar quadrotor dynamical system. $N = 50$ support states are used, time span $T = 4.0\sim 6.0$, radius $r+\epsilon_{sdf} = 1.5$, $\Sigma_{obs} = 6.0I\sim7.5I$, low temperature $\hat{T}_{low} = 1.0$, high temperature $\hat{T}_{high} = 5.0$. 
  }
  \label{fig:ltv_4exp}
\end{figure*}

{\em (c) More experiments with different settings.}
We conducted experiments under four different settings with obstacles in the environment. Figure~\ref{fig:ltv_4exp} shows the results of the nonlinear system of the planar quadrotor \eqref{eq:planar_quad} for the $4$ experiment settings. 

\begin{figure}[ht]
\centering
    \begin{subfigure}[ht]{0.45\linewidth}
    \centering
    \includegraphics[width=0.95\linewidth]{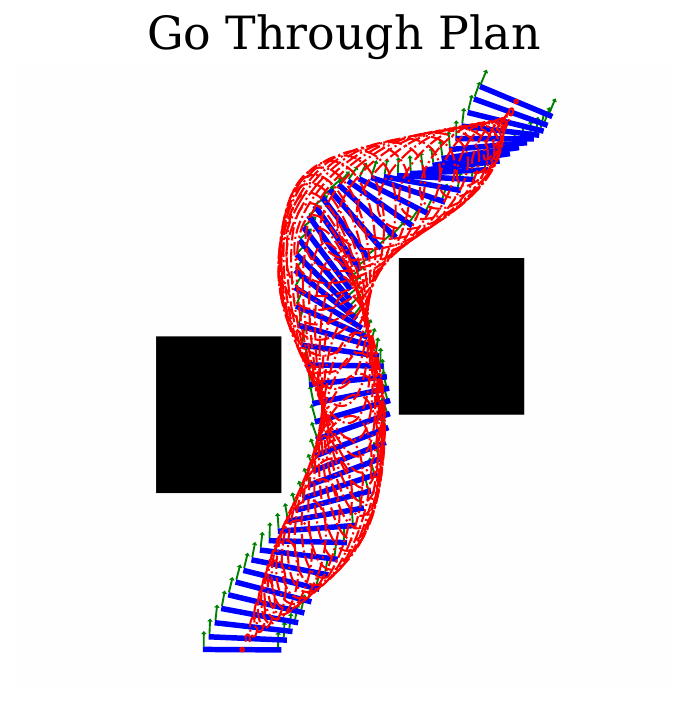}
    \end{subfigure}
    \begin{subfigure}[ht]{0.45\linewidth}
    \centering
    \includegraphics[width=0.95\linewidth]{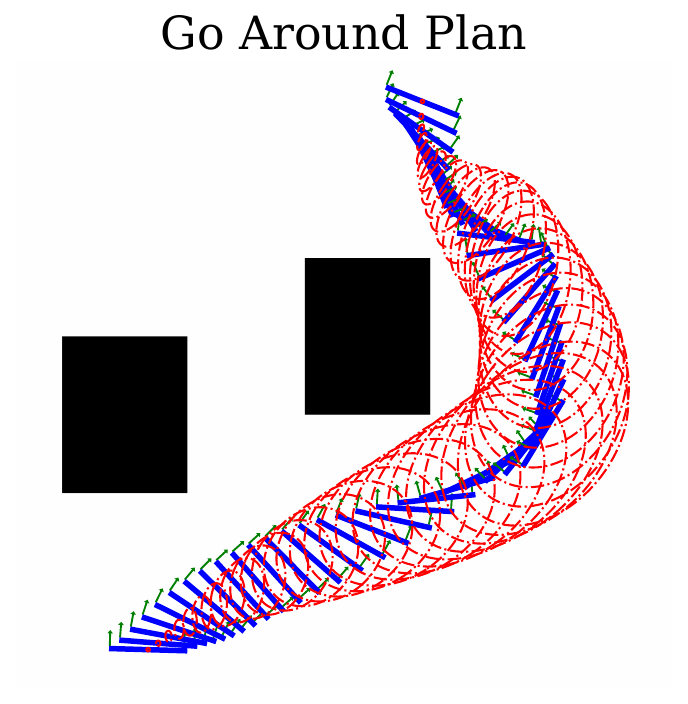}
    \end{subfigure}
  \caption{ Entropy regularized robust planning in a narrow gap environment.}
  \label{fig:ltv_narrow_gap}
\end{figure}

{\em (d) Go through or go around a narrow gap? Robust motion planning through entropy regularization.}
As motion planning with obstacles is a non-convex problem \cite{yu2021convex}, and the solution is multi-modal. Risk and robustness of the solution are considered in our formulation by introducing the entropy of $q_\theta$ into the objective. This section shows an experiment for the planar quadrotor to fly through a narrow gap \cite{yu2023gaussian}. Two motion plan modes are obtained, as shown in Figure~\ref{fig:ltv_narrow_gap}. One plan (go-through) is visually riskier than the other (go-around). 

Our formulation provides a quantitative metric for comparing the optimality and robustness of the two plans through the entropy cost. Table~\ref{tab:plans_cost_comparison} reports the sum of the prior and collision costs, the entropy costs, and the total costs. By incorporating lower entropy costs, the proposed method favors safer plans over shorter but riskier ones.

\begin{table}[th]
\centering
\resizebox{\columnwidth}{!}{
\begin{tabular}{|c|c|c|c|c|c|}
\hline
 & Prior & Collision & MP & Entropy & Total \\ \hline
Go Through &  \textbf{138.96}  &  1.26  &  \textbf{140.22} & 1232.36  &  1372.58  \\ \hline
Go Around  &  149.34  &  \textbf{0.0369}  & 149.38 &\textbf{1163.22}  &  \textbf{1312.6}  \\ \hline
\end{tabular}
}
\caption{Comparing costs for two plans in Fig.\ref{fig:ltv_narrow_gap}. The `MP' is short for the sum of prior and collision costs, and entropy represents the robustness. }
\label{tab:plans_cost_comparison}
\end{table}

\section{Conclusion}\label{sec:conclusion}
This work describes a distributed Gaussian Variational Inference approach to motion planning under uncertainties. The optimal trajectory distribution to a stochastic control problem serves as the target posterior in a variational inference paradigm. We leveraged this inference's underlying sparse factor graph structure and proposed a distributed computation framework to solve the VI problem in parallel on GPU. Numerical experiments show the effectiveness of the proposed methods on an LTV system, and comparison studies demonstrated the computational efficiency.

\appendix
\section{Proof of Theorem \ref{thm:KLproximal_updates}}
\label{sec:appendix_A}
\begin{proof}
    Taking the gradient of the objective in \eqref{eq:prox-KL-optimization} with respect to 
$\mu_{\theta}$ and setting it to zero, we obtain
\begin{equation*}
    \label{eq:optimality_prox_KL_mu}
    0 = \nabla_{\mu_\theta} \psi_d(\theta_k) + \bK^{-1}(\mu_{\theta} - \bmu) + \frac{1}{\beta_k}\Sigma_{\theta_k}^{-1}(\mu_{\theta} - \mu_{\theta_k}),
\end{equation*}
which yields the update for the mean
\begin{equation*}
\label{eq:mu_update}
( \bK^{-1} + \frac{1}{\beta_k} \Sigma_{\theta_k}^{-1} ) \mu_{\theta_{k+1}} = -\nabla_{\mu_\theta} \psi_d(\theta_k) + \bK^{-1}\bmu + \frac{1}{\beta_k}\Sigma_{\theta_k}^{-1}\mu_{\theta_k}.
\end{equation*}
Since $\nabla_{\Sigma_{\theta}^{-1}} (\cdot) = -\Sigma_{\theta}\,\nabla_{\Sigma_{\theta}}(\cdot)\,\Sigma_{\theta}$, the necessary condition for updating $\Sigma_\theta^{-1}$ follows by computing the gradient of the objective in \eqref{eq:prox-KL-optimization} with respect to $\Sigma_{\theta}$, setting it to zero, and obtaining
\begin{align*}
    \label{eq:optimality_prox_KL_Sigma}
    0 
    &= \nabla_{\Sigma_\theta} \psi_d(\theta_k) + \frac{1}{2}(\frac{1}{\beta_k} \Sigma_{\theta_k}^{-1} + \bK^{-1} - (1+\frac{1}{\beta_k})\Sigma_{\theta_{k+1}}^{-1}),
\end{align*}
from which the update rule for $\Sigma_{\theta}^{-1}$ follows
\begin{equation*}
\label{eq:Sigma_inv_update}
\Sigma_{\theta_{k+1}}^{-1} = \frac{\beta_k}{\beta_k + 1} \left( 2 \nabla_{\Sigma_{\theta}} \psi_d(\theta_k) + \bK^{-1} + \frac{1}{\beta_k} \Sigma_{\theta_k}^{-1} \right).
\end{equation*}
\end{proof}

\section{Theoretical Analysis of P-GVIMP}
\label{sec:pgvimp_analysis}
To establish the approximation accuracy and convergence guarantees of the proposed P-GVIMP algorithm, we define $\operatorname{KL}\left(\theta_{k+1}\! \parallel\! \theta_{k}\right) := \operatorname{KL} \left(q(x \!\mid\! \theta_{k+1})\! \parallel \! q\left(x\! \mid\! \theta_{k}\right)\right)$ and make the following assumptions:

\textbf{(A1)}: The ELBO $\mathcal{L}$ is continuous and admits a maximum.

\textbf{(A2)}: The function $\psi_d$ is L-smooth in $\mathcal{S}$
\begin{equation*}
    \left\|\nabla \psi_d(\boldsymbol{\theta})-\nabla \psi_d\left(\boldsymbol{\theta}^{\prime}\right)\right\| \leq L\left\|\boldsymbol{\theta}-\boldsymbol{\theta}^{\prime}\right\|, \forall \boldsymbol{\theta}, \boldsymbol{\theta}^{\prime} \in \mathcal{S}
\end{equation*}

\textbf{(A3)}: The function $\psi_e$ is convex.

\textbf{(A4)}: There exists an $\alpha >0$ such that
\begin{equation*}
    \left(\theta_{k+1}-\theta_{k}\right)^{T} \nabla \operatorname{KL}\left(\theta_{k+1}\! \parallel \!\theta_{k}\right) \!\geq \alpha \left\|\theta_{k+1}-\theta_{k}\right\|^{2}
\end{equation*}

\subsection*{\textbf{Approximation Error Bound}}
\begin{proof}
    Let $\delta:=\theta - \theta_k$, and define the first-order error
    \begin{equation*}
        \varepsilon(\theta)=\psi_{d}(\theta)-\left[\psi_{d}\left(\theta_{k}\right)+\nabla \psi_{d}\left(\theta_{k}\right)^T\delta\right] .
    \end{equation*}
    By the $L$-smoothness assumption on $\psi_d$ \textbf{(A2)},
    \begin{equation*}
        \psi_{d}\left(\theta\right) \leq \psi_{d}\left(\theta_{k}\right)+\nabla \psi_{d}\left(\theta_{k}\right)^T (\theta-\theta_k)+\frac{L}{2}\left\|\theta-\theta_k\right\|^{2}
    \end{equation*}
    Similarly, since $-\psi_d$ is also $L$-smooth,
    \begin{equation*}
        -\psi_{d}\left(\theta\right) \leq -\psi_{d}\left(\theta_{k}\right)-\nabla \psi_{d}\left(\theta_{k}\right)^T (\theta-\theta_k)+\frac{L}{2}\left\|\theta-\theta_k\right\|^{2}
    \end{equation*}
    Combining the two bounds gives
    \begin{equation*}
        |\varepsilon(\theta)| \leq \frac{L}{2}\|\delta\|^{2} \quad \Longrightarrow \quad \varepsilon(\theta)=\mathcal{O}\left(\|\delta\|^{2}\right) .
    \end{equation*}
    which shows that the first-order approximation error is of order $\mathcal{O}(\|\delta\|^{2})$ over any region where $\psi_d$ is $L$-smooth.
\end{proof}

\subsection*{\textbf{Convergence Guarantee}}
\begin{proof}
    Given \textbf{(A3)} and \textbf{(A4)}, Lemma~1 in \cite{khan2015faster} yields
    \begin{equation*}
        \left(\theta_{k}-\theta_{k+1}\right)^{T} \nabla \psi_d(\theta_k) \geq \frac{\alpha}{\beta}\left\|\theta_{k+1}-\theta_{k}\right\|^{2}+\psi_e\left(\theta_{k+1}\right)-\psi_e\left(\theta_{k}\right).
    \end{equation*}
    The $L$-smoothness assumption \textbf{(A2)} gives
    \begin{equation*}
        \psi_d\left(\theta_{k+1}\right) \leq \psi_d\left(\theta_{k}\right)+\nabla \psi_d\left(\theta_{k}\right)^T\left(\theta_{k+1}-\theta_{k}\right)+\frac{L}{2}\left\|\theta_{k+1}-\theta_{k}\right\|^{2}
    \end{equation*}
    Combining the above two inequalities
    \begin{align*}
        \psi_d(\theta_{k+1}) + \psi_e(\theta_{k+1}) 
        &\leq 
        \psi_d(\theta_k) + \psi_e(\theta_k) 
        - \left(\frac{\alpha}{\beta} - \frac{L}{2}\right)\|\theta_{k+1}-\theta_k\|^{2} \\[-1pt]
        \Rightarrow\quad
        \mathcal{L}(\theta_{k+1}) 
        &\leq 
        \mathcal{L}(\theta_k) 
        - \left(\frac{\alpha}{\beta} - \frac{L}{2}\right)\|\theta_{k+1}-\theta_k\|^{2}.
    \end{align*}
    Therefore, for any step size $0<\beta<\frac{2\alpha}{L}$, the sequence $\{\mathcal{L}(\theta_k)\}$ is strictly increasing and bounded by its maximum. Consequently, $\left( \frac{\alpha}{\beta} - \frac{L}{2} \right) \left\|\theta_{k+1}-\theta_{k}\right\|^{2} \to 0$, and hence $\left\|\theta_{k+1}-\theta_{k}\right\| \to 0$.
\end{proof}

\section{Implementation Details}
This section presents the implementation details of the P-GVIMP Algorithm \ref{alg:distributed-gvimp}. \paragraph{Step size selection}
The proximal point algorithm does not guarantee that the updated distribution at each step remains inside the distribution family $\mathcal{Q}$, especially when the step sizes $\beta_k$ are too large. The updates do not guarantee that the updated covariances are always positive definite. To enhance convergence efficiency, we control $\beta_k$ by imposing an upper bound on the KL divergence between the updated and current distributions. To this end, we solve the convex program
\begin{equation}
    \begin{aligned}
        & \beta_k^\star \ = \ \max_{\beta_k > 0} \  \beta_k
        \\
        & \text{s.t.} \quad {\rm KL}\left(q_{\theta_{k+1}}(\mathbf{x}) \| q_{\theta_k}(\mathbf{x})\right) \leq \epsilon, \; \Sigma^{-1}_{\theta_{k+1}} \succ 0.
    \end{aligned}
\end{equation} 
In practice, a bi-section algorithm is used to choose a step size that keeps the next step distribution close to the current one. 

\bibliographystyle{elsarticle-num}

\bibliography{root}

\begin{thebibliography}{10}
\expandafter\ifx\csname url\endcsname\relax
  \def\url#1{\texttt{#1}}\fi
\expandafter\ifx\csname urlprefix\endcsname\relax\def\urlprefix{URL }\fi
\expandafter\ifx\csname href\endcsname\relax
  \def\href#1#2{#2} \def\path#1{#1}\fi

\bibitem{lavalle2006planning}
S.~M. LaValle, Planning algorithms, Cambridge university press, 2006.

\bibitem{gonzalez2015review}
D.~Gonz{\'a}lez, J.~P{\'e}rez, V.~Milan{\'e}s, F.~Nashashibi, A review of motion planning techniques for automated vehicles, IEEE Transactions on intelligent transportation systems 17~(4) (2015) 1135--1145.

\bibitem{ratliff2009chomp}
N.~Ratliff, M.~Zucker, J.~A. Bagnell, S.~Srinivasa, Chomp: Gradient optimization techniques for efficient motion planning, in: IEEE international conference on robotics and automation, 2009, pp. 489--494.

\bibitem{schulman2014motion}
J.~Schulman, Y.~Duan, J.~Ho, A.~Lee, I.~Awwal, H.~Bradlow, J.~Pan, S.~Patil, K.~Goldberg, P.~Abbeel, Motion planning with sequential convex optimization and convex collision checking, The International Journal of Robotics Research 33~(9) (2014) 1251--1270.

\bibitem{chen2022should}
G.~Chen, H.~Yu, W.~Dong, X.~Sheng, X.~Zhu, H.~Ding, What should be the input: Investigating the environment representations in sim-to-real transfer for navigation tasks, Robotics and Autonomous Systems 153 (2022) 104081.

\bibitem{chen2000nonlinear}
W.-H. Chen, D.~J. Ballance, P.~J. Gawthrop, J.~O'Reilly, A nonlinear disturbance observer for robotic manipulators, IEEE Transactions on industrial Electronics 47~(4) (2000) 932--938.

\bibitem{kalakrishnan2011stomp}
M.~Kalakrishnan, S.~Chitta, E.~Theodorou, P.~Pastor, S.~Schaal, Stomp: Stochastic trajectory optimization for motion planning, in: IEEE international conference on robotics and automation, 2011, pp. 4569--4574.

\bibitem{aastrom2012introduction}
K.~J. {\AA}str{\"o}m, Introduction to stochastic control theory, Courier Corporation, 2012.

\bibitem{thrun2002probabilistic}
S.~Thrun, Probabilistic robotics, Communications of the ACM 45~(3) (2002) 52--57.

\bibitem{hoshino2025path}
K.~Hoshino, H.~Yu, T.~Tanaka, Y.~Chen, Path integral control of partially observed systems via fully observable control approximations, Systems \& Control Letters 204 (2025) 106185.

\bibitem{mukadam2016gaussian}
M.~Mukadam, X.~Yan, B.~Boots, Gaussian process motion planning, in: IEEE international conference on robotics and automation (ICRA), 2016, pp. 9--15.

\bibitem{mukadam2018continuous}
M.~Mukadam, J.~Dong, X.~Yan, F.~Dellaert, B.~Boots, Continuous-time gaussian process motion planning via probabilistic inference, The International Journal of Robotics Research 37~(11) (2018) 1319--1340.

\bibitem{hoffman2013stochastic}
M.~D. Hoffman, D.~M. Blei, C.~Wang, J.~Paisley, Stochastic variational inference, Journal of Machine Learning Research (2013).

\bibitem{blei2017variational}
D.~M. Blei, A.~Kucukelbir, J.~D. McAuliffe, Variational inference: A review for statisticians, Journal of the American Statistical Association 112~(518) (2017) 859--877.

\bibitem{yu2023gaussian}
H.~Yu, Y.~Chen, A gaussian variational inference approach to motion planning, IEEE Robotics and Automation Letters 8~(5) (2023) 2518--2525.

\bibitem{yu2023stochastic}
H.~Yu, Y.~Chen, Stochastic motion planning as gaussian variational inference: Theory and algorithms, arXiv preprint arXiv:2308.14985 (2023).

\bibitem{cosier2024unifying}
L.~C. Cosier, R.~Iordan, S.~N. Zwane, G.~Franzese, J.~T. Wilson, M.~Deisenroth, A.~Terenin, Y.~Bekiroglu, A unifying variational framework for gaussian process motion planning, in: International Conference on Artificial Intelligence and Statistics, PMLR, 2024, pp. 1315--1323.

\bibitem{power2024constrained}
T.~Power, D.~Berenson, Constrained stein variational trajectory optimization, IEEE Transactions on Robotics (2024).

\bibitem{hsu1998finding}
D.~Hsu, L.~E. Kavraki, J.-C. Latombe, R.~Motwani, S.~Sorkin, et~al., On finding narrow passages with probabilistic roadmap planners, in: Robotics: the algorithmic perspective: 1998 workshop on the algorithmic foundations of robotics, 1998, pp. 141--154.

\bibitem{chen2016relation}
Y.~Chen, T.~T. Georgiou, M.~Pavon, On the relation between optimal transport and schr{\"o}dinger bridges: A stochastic control viewpoint, Journal of Optimization Theory and Applications 169 (2016) 671--691.

\bibitem{chen2016optimal}
Y.~Chen, T.~T. Georgiou, M.~Pavon, Optimal transport over a linear dynamical system, IEEE Transactions on Automatic Control 62~(5) (2016) 2137--2152.

\bibitem{haarnoja2018soft}
T.~Haarnoja, A.~Zhou, P.~Abbeel, S.~Levine, Soft actor-critic: Off-policy maximum entropy deep reinforcement learning with a stochastic actor, in: International conference on machine learning, Pmlr, 2018, pp. 1861--1870.

\bibitem{zhao2019maximum}
R.~Zhao, X.~Sun, V.~Tresp, Maximum entropy-regularized multi-goal reinforcement learning, in: International Conference on Machine Learning, PMLR, 2019, pp. 7553--7562.

\bibitem{barfoot2014batch}
T.~D. Barfoot, C.~H. Tong, S.~S{\"a}rkk{\"a}, Batch continuous-time trajectory estimation as exactly sparse gaussian process regression., in: Robotics: Science and Systems, Vol.~10, Citeseer, 2014, pp. 1--10.

\bibitem{barfoot2020exactly}
T.~D. Barfoot, J.~R. Forbes, D.~J. Yoon, Exactly sparse gaussian variational inference with application to derivative-free batch nonlinear state estimation, The International Journal of Robotics Research 39~(13) (2020) 1473--1502.

\bibitem{khan2015kullback}
M.~E.~E. Khan, P.~Baqu{\'e}, F.~Fleuret, P.~Fua, Kullback-leibler proximal variational inference, Advances in neural information processing systems 28 (2015).

\bibitem{khan2015faster}
M.~E. Khan, R.~Babanezhad, W.~Lin, M.~Schmidt, M.~Sugiyama, Faster stochastic variational inference using proximal-gradient methods with general divergence functions, in: Proceedings of the Conference on Uncertainty in Artificial Intelligence, 2016, pp. 319--328.

\bibitem{shental2008gaussian}
O.~Shental, P.~H. Siegel, J.~K. Wolf, D.~Bickson, D.~Dolev, Gaussian belief propagation solver for systems of linear equations, in: IEEE International Symposium on Information Theory, 2008, pp. 1863--1867.

\bibitem{osa2020multimodal}
T.~Osa, Multimodal trajectory optimization for motion planning, The International Journal of Robotics Research 39~(8) (2020) 983--1001.

\bibitem{carvalho2023motion}
J.~Carvalho, A.~T. Le, M.~Baierl, D.~Koert, J.~Peters, Motion planning diffusion: Learning and planning of robot motions with diffusion models, in: 2023 IEEE/RSJ International Conference on Intelligent Robots and Systems (IROS), IEEE, 2023, pp. 1916--1923.

\bibitem{sarkka2019applied}
S.~S{\"a}rkk{\"a}, A.~Solin, Applied stochastic differential equations, Vol.~10, Cambridge University Press, 2019.

\bibitem{dellaert2021factor}
F.~Dellaert, Factor graphs: Exploiting structure in robotics, Annual Review of Control, Robotics, and Autonomous Systems 4~(1) (2021) 141--166.

\bibitem{barfoot2024state}
T.~D. Barfoot, State estimation for robotics, Cambridge University Press, 2024.

\bibitem{seo2024sequential}
M.-W. Seo, S.~S. Kia, Sequential gaussian variational inference for nonlinear state estimation and its application in robot navigation, IEEE Robotics and Automation Letters (2024).

\bibitem{diao2023forward}
M.~Z. Diao, K.~Balasubramanian, S.~Chewi, A.~Salim, Forward-backward gaussian variational inference via jko in the bures-wasserstein space, in: International Conference on Machine Learning, PMLR, 2023, pp. 7960--7991.

\bibitem{gelb1974applied}
A.~Gelb, et~al., Applied optimal estimation, MIT press, 1974.

\bibitem{julier2004unscented}
S.~J. Julier, J.~K. Uhlmann, Unscented filtering and nonlinear estimation, Proceedings of the IEEE 92~(3) (2004) 401--422.

\bibitem{pan2012collision}
J.~Pan, L.~Zhang, D.~Manocha, Collision-free and smooth trajectory computation in cluttered environments, The International Journal of Robotics Research 31~(10) (2012) 1155--1175.

\bibitem{pan2012gpu}
J.~Pan, D.~Manocha, Gpu-based parallel collision detection for fast motion planning, The International Journal of Robotics Research 31~(2) (2012) 187--200.

\bibitem{janson2015fast}
L.~Janson, E.~Schmerling, A.~Clark, M.~Pavone, Fast marching tree: A fast marching sampling-based method for optimal motion planning in many dimensions, The International journal of robotics research 34~(7) (2015) 883--921.

\bibitem{gammell2015batch}
J.~D. Gammell, S.~S. Srinivasa, T.~D. Barfoot, Batch informed trees (bit*): Sampling-based optimal planning via the heuristically guided search of implicit random geometric graphs, in: 2015 IEEE international conference on robotics and automation (ICRA), IEEE, 2015, pp. 3067--3074.

\bibitem{van2011reciprocal}
J.~Van Den~Berg, S.~J. Guy, M.~Lin, D.~Manocha, Reciprocal n-body collision avoidance, in: Robotics Research: The 14th International Symposium ISRR, Springer, 2011, pp. 3--19.

\bibitem{sharon2015conflict}
G.~Sharon, R.~Stern, A.~Felner, N.~R. Sturtevant, Conflict-based search for optimal multi-agent pathfinding, Artificial intelligence 219 (2015) 40--66.

\bibitem{huang2025prrtc}
C.~H. Huang, P.~Jadhav, B.~Plancher, Z.~Kingston, prrtc: Gpu-parallel rrt-connect for fast, consistent, and low-cost motion planning, arXiv preprint arXiv:2503.06757 (2025).

\bibitem{yu2024optimal}
H.~Yu, D.~F. Franco, A.~M. Johnson, Y.~Chen, Optimal covariance steering of linear stochastic systems with hybrid transitions, arXiv preprint arXiv:2410.13222 (2024).

\bibitem{parikh2014proximal}
N.~Parikh, S.~Boyd, et~al., Proximal algorithms, Foundations and trends{\textregistered} in Optimization 1~(3) (2014) 127--239.

\bibitem{chretien2000kullback}
S.~Chr{\'e}tien, A.~O. Hero, Kullback proximal algorithms for maximum-likelihood estimation, IEEE transactions on information theory 46~(5) (2000) 1800--1810.

\bibitem{gerstner1998numerical}
T.~Gerstner, M.~Griebel, Numerical integration using sparse grids, Numerical algorithms 18~(3) (1998) 209--232.

\bibitem{heiss2008likelihood}
F.~Heiss, V.~Winschel, Likelihood approximation by numerical integration on sparse grids, Journal of Econometrics 144~(1) (2008) 62--80.

\bibitem{Ortiz2021visualGBP}
J.~Ortiz, T.~Evans, A.~J. Davison, A visual introduction to gaussian belief propagation, arXiv preprint arXiv:2107.02308 (2021).

\bibitem{arasaratnam2007discrete}
I.~Arasaratnam, S.~Haykin, R.~J. Elliott, Discrete-time nonlinear filtering algorithms using gauss--hermite quadrature, Proceedings of the IEEE 95~(5) (2007) 953--977.

\bibitem{rooks2006harmonious}
B.~Rooks, The harmonious robot, Industrial Robot: An International Journal 33~(2) (2006) 125--130.

\bibitem{garage2012pr2}
W.~Garage, Pr2 user manual (2012).

\bibitem{quigley2009ros}
M.~Quigley, K.~Conley, B.~Gerkey, J.~Faust, T.~Foote, J.~Leibs, R.~Wheeler, A.~Y. Ng, et~al., Ros: an open-source robot operating system, in: ICRA workshop on open source software, Vol.~3, Kobe, 2009, p.~5.

\bibitem{coleman2014reducing}
D.~Coleman, I.~Sucan, S.~Chitta, N.~Correll, Reducing the barrier to entry of complex robotic software: a moveit! case study, arXiv preprint arXiv:1404.3785 (2014).

\bibitem{yu2021convex}
H.~Yu, J.~Moyalan, D.~Tellez-Castro, U.~Vaidya, Y.~Chen, Convex optimal control synthesis under safety constraints, in: 2021 60th IEEE Conference on Decision and Control (CDC), IEEE, 2021, pp. 4615--4621.

\end{thebibliography}

\end{document}